\documentclass{ecai}

\usepackage{latexsym}
\usepackage{amssymb}
\usepackage{amsmath}
\usepackage{amsthm}
\usepackage{booktabs}
\usepackage{enumitem}
\usepackage{graphicx}
\usepackage{color}
\usepackage{booktabs} %
\usepackage{amsfonts}
\usepackage{dsfont}
\usepackage{multirow}
\usepackage{hyperref}
\usepackage[numbers]{natbib}

\newcommand{\BibTeX}{B\kern-.05em{\sc i\kern-.025em b}\kern-.08em\TeX}

\begin{document}

\begin{frontmatter}

\title{Efficient Model-Stealing Attacks Against Inductive Graph Neural Networks}

\author[A,B]{\fnms{Marcin}~\snm{Podhajski}\thanks{Corresponding Author. Email: marcin.podhajski@ideas-ncbr.pl}}
\author[A,C]{\fnms{Jan}~\snm{Dubiński}}
\author[D]{\fnms{Franziska}~\snm{Boenisch}}
\author[D]{\fnms{Adam}~\snm{Dziedzic}}
\author[B]{\fnms{\\Agnieszka}~\snm{Pregowska}}
\author[A,E]{\fnms{Tomasz P.}~\snm{Michalak}} 

\address[A]{IDEAS NCBR, {\small 00-801 Warsaw, Poland}}
\address[B]{Institute of Fundamental Technological Research, Polish Academy of Sciences, {\small Pawińskiego 5B, 02-106 Warsaw, Poland}}
\address[C]{Warsaw University of Technology, Faculty of Electronics and Information Technology, {\small 00-661 Warsaw, Poland}}
\address[D]{CISPA, {\small 66123 Saarbrücken, Germany}}
\address[E]{University of Warsaw, Faculty of Mathematics, Informatics and Mechanics, {\small 00-927 Warsaw, Poland}}

\begin{abstract}
\setcounter{footnote}{1}
Graph Neural Networks (GNNs) are recognized as potent tools for processing real-world data organized in graph structures. Especially inductive GNNs, which allow for the processing of graph-structured data without relying on predefined graph structures, are becoming increasingly important in a wide range of applications. As such these networks become attractive targets for model-stealing attacks where an adversary seeks to replicate the functionality of the targeted network. Significant efforts have been devoted to developing model-stealing attacks that extract models trained on images and texts. However, little attention has been given to stealing GNNs trained on graph data. This paper identifies a new method of performing \textit{unsupervised} model-stealing attacks against inductive GNNs, utilizing graph contrastive learning and spectral graph augmentations to efficiently extract information from the targeted model. The new type of attack is thoroughly evaluated on six datasets and %
the results show that our approach outperforms the current state-of-the-art by Shen et al. (2021). In particular, our attack surpasses the baseline across all benchmarks, attaining superior fidelity and downstream accuracy of the stolen model while necessitating fewer queries directed toward the target model.\footnotemark[1]
\end{abstract}
\end{frontmatter}

\section{Introduction}
\par
\footnotetext[1]{Code available at \url{https://github.com/m-podhajski/EfficientGNNStealing}}
Graph Neural Networks (GNNs) are specifically designed to operate on graph-structured data, such as molecules, social networks or complex infrastructure \cite{10.1145/3308558.3313488,doi:10.1021/acs.jcim.9b00237, powergnn}. Unlike models developed for images or text, GNNs simultaneously consider both node features and graph structures. GNNs have garnered significant attention for their effectiveness in tasks involving structured data. A major advantage of GNNs is their ability to scale to large graphs while preserving computational efficiency. However, it is also well-documented that Machine Learning (ML) models, including GNNs, are vulnerable to \emph{model stealing attacks}~\cite{TZJRR16,JCBKP20, pmlr-v162-dziedzic22a}, where an adversary with query access to a \textit{target (victim) model} can extract its parameters or functionality, often at a fraction of the cost required to train the model from scratch.

In such scenarios, the adversary sends a series of queries as inputs to the target model's API and obtains the corresponding outputs. A local surrogate model is then trained to mimic the outputs of the target model for the same input data. Consequently, the surrogate model not only compromises the intellectual property of the target model but also potentially enables further attacks, such as extracting private training data~\cite{LWHSZBCFZ22} or constructing adversarial examples~\cite{PMGJCS17}.

\begin{figure}[t]
\centering
\includegraphics[trim={0cm -0.1cm 0cm 0cm},clip, width=1\linewidth]{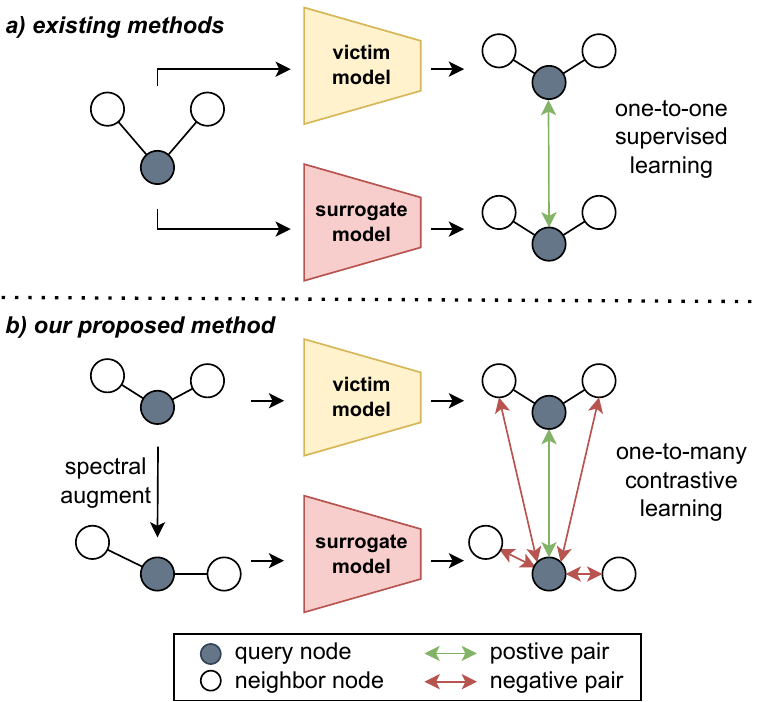} 
\caption{Existing model stealing methods rely on aligning only the predictions for individual graph nodes sent to the victim and surrogate model. The new method identified in this paper greatly increases the amount of information extracted from the victim model outputs. We align the node representation from the surrogate model with the corresponding representation from the victim model while simultaneously distinguishing it from the representation of other nodes. By integrating spectral augmentations to augment the graph inputs of the surrogate models during its training process, we further increase the efficiency of the stealing process per query sent to the victim model.}
\vspace{0.5cm}
\label{fig:teaser}
\end{figure}

\par
It is worth stressing that the majority of the current efforts on model stealing attacks concentrate on ML models for images and text data~\cite{TZJRR16,OSF19,WG18,KTPPI20}. 
Despite the growing significance of GNNs, model-stealing attacks against these types of neural networks are severely underexplored compared to image or text models. To our knowledge, only one study~\cite{shen2021model} addresses attacks on inductive GNNs, i.e. a type of model that can generalize well to unseen nodes. The authors define the threat model for model stealing attacks on inductive GNNs and describe six attack methods based on the adversary's background knowledge and the responses of the target models. During the attack, the response of the surrogate model is aligned with the response of the target model by minimizing the RMSE loss between them. Although their approach yields promising results, the underlying method of training the surrogate model does not efficiently utilize the information available from the target's responses. This increases the number of queries required to extract the target model's API which results in higher chances of attack detection and mitigation.

Inspired by recent approaches to encoder model stealing \cite{pmlr-v162-dziedzic22a, liu2022stolenencoder, sha2023cant} we leverage the significant fact that augmented graph inputs should produce similar model outputs. Thus, the GNN model-stealing attack identified in this paper utilizes a contrastive objective at the node level. Specifically, for an input graph, we generate two graph views---one from the target model and one from the surrogate model for an augmented graph input. The surrogate model is trained by maximizing the agreement of node representations in these two views. This contrastive objective considers both positive and negative pairs between the node representations from the target and surrogate model. To augment the graph input to the surrogate model we leverage graph transformation operations derived via graph spectral analysis, enabling contrastive learning to capture useful structural representations. Those transformations include spectral graph cropping and graph frequency components reordering. Thus, the proposed methods that combine contrastive learning loss and effective graph-specific augmentations enable efficient usage of the information from the target model outputs. As a result, our approach fits into the framework established by Shen et al. (2021) \cite{shen2021model}. It presents an enhanced surrogate model training methodology that significantly augments the efficiency and performance of the model-stealing process, outperforming the state-of-the-art method with less than half of its query budget.
\par
We evaluate all proposed attacks on three popular inductive GNN models including GraphSAGE~\cite{HYL17}, Graph Attention Network (GAT)~\cite{VCCRLB18}, and Graph Isomorphism Network (GIN)~\cite{XHLJ19} with six benchmark datasets, i.e., DBLP~\cite{PWZZW16}, Pubmed~\cite{SNBGGE08}, Citeseer Full~\cite{GBL98}, Coauthor Physics~\cite{SMBG18}, ACM~\cite{WJSWYCY19}, and Amazon Co-purchase Network for Photos~\cite{MTSH15}. The experiments demonstrate that our model-stealing attack consistently outperforms the state-of-the-art baseline approach across the vast majority of datasets and models. In the particular case of using SAGE as the surrogate model the increase in accuracy reached 36.8 \%pt. for a GIN target trained on Amazon Co-purchase Network for Photos.
\par
Overall, the main contributions of this work are as follows:
\begin{itemize}
    \item We identify a new unsupervised GNN model stealing attack trained with a contrastive objective at the node level that compels the embedding of corresponding nodes from the surrogate and victim models to align while also distinguishing them from embeddings of other nodes within the surrogate and the target model.
    \item We adapt spectral graph augmentations to generate multiple graph views for the surrogate model in the contrastive learning process.
    \item We perform an extensive comparison of the newly identified stealing technique to the state of the art. The new stealing process demonstrates better performance while requiring fewer queries sent to the target model.
\end{itemize}

\section{Preliminaries}\label{sec:preliminaries}

\begin{table}[h]
\centering
\caption{\small Notations used in the presented paper. Lowercase letters denote scalars, bold lowercase letters denote vectors and bold uppercase letters denote matrices.}

\vspace{0.5cm}
\scalebox{1}{
\begin{tabular}{l|l}
 \toprule 
 \textbf{Notation} & \textbf{Description} \\ \midrule
    $\mathbf{G}=(\mathbf{V}, \mathbf{E}, \mathbf{X}, \mathbf{C})$   & graph   \\
    $v,u \in V$ &  node \\
    $n=|\mathbf{V}|$ & number of nodes \\
    $d \in \mathbb{N}$ & dimension of a node feature vector \\
    $b \in \mathbb{N}$ & dimension of a node embedding vector \\
    $\mathbf{A} \in \{0,1\} ^ {n \times n}$ &  adjacency matrix \\ 
    $\mathbf{A'} \in \langle-1,1\rangle ^ {n \times n}$ &  augmented adjacency matrix \\ 
    $\mathbf{X} \in \mathbb{R} ^ {n \times d}$ & feature matrix \\ 
    $\mathbf{H} \in \mathbb{R}^{n \times b} $ & embedding matrix \\
    $\mathbf{h}_v^i \in \mathbb{R}^{ b} $ & node embedding at layer $i$ \\
    $\mathcal{N}(v)$ & neighborhood of $v$  \\
    $\mathcal{M}_T$/$\mathcal{M}_S$ & target/surrogate GNN model \\
 \bottomrule	 
\end{tabular}
}
\label{tab:notations}
\end{table}

Graph Neural Networks are a class of neural networks that use the graph structure $\mathbf{A}$ as well as node features  $\mathbf{X}$ as the input to the model. GNNs learn an embedding vector of a node $h_v$ or the entire graph $H$. These embeddings can be then used in a range of important tasks, including node classification ~\cite{HYL17, KW17}, link prediction ~\cite{PAS14, GL16, BKW17} or graphs and subgraph classification ~\cite{NEURIPS2020_5bca8566, LLK19, YYMRHL18}.
\par
Most of the modern GNNs follow a neighborhood aggregation strategy. We start with 
$\mathbf{h}_v^0=\mathbf{X}_v$ and after $k$ iterations of aggregation, the $k$-th layer of a GNN is $$\mathbf{h}_v^k=AGGREGATE(\mathbf{h}_v^{k-1},MSG(\mathbf{h}_v^{k-1},\mathbf{h}_u^{k-1}),$$ $u \in \mathcal{N}(v) $. In effect, the node embedding at the $k$-th layer captures the information within its $k$-hop neighborhood.

\par
The training of GNNs occurs in two distinct settings: the transductive setting and the inductive setting. In the transductive setting the data consists of a fixed graph and a subset of nodes is assigned labels and utilized during the training process, while another subset remains unlabeled. The primary objective is to leverage the labeled nodes and predict the labels for the previously unlabeled nodes within the same graph. However, the transductive models do not generalize to newly introduced nodes.
\par
In the inductive setting, the GNN is designed to generalize its learning to accommodate new, unseen nodes or graphs beyond those encountered during training. 
The first model capable of inductive learning was GraphSAGE introduced by \cite{HYL17}. Other inductive models include Graph Attention Network (GAT) proposed by \cite{VCCRLB18}, and Graph Isomorphism Network (GIN) by \cite{XHLJ19}.

\section{Threat Model}\label{sec:threat:model}
This section outlines the threat model, detailing the attack setting, adversary's goals, and capabilities. The proposed approach is based on the threat model widely presented in \cite{shen2021model}.

\textbf{Attack Setting.}
We investigate a challenging \textit{black-box} scenario where the adversary is entirely unaware of the target GNN model's specifics, including its parameters and architecture, and cannot interfere with the training process or the training graph $\mathbf{G}_O$. Our study focuses on the responses to three distinct types of node-level queries. In this context, the target model is an inductive GNN that utilizes the $l$-hop subgraph $\mathbf{G}_v^l$ of a node $v$ as input to generate outputs for that node. These outputs could be in the form of predicted posterior probabilities, node embedding vectors, or a 2-dimensional t-SNE projection.

\textbf{Adversary's Goal.}
According to the classification established by \cite{JCBKP20}, adversaries typically have two primary objectives: theft or reconnaissance. A \emph{theft adversary} aims to create a surrogate model $\mathcal{M}_S$ that replicates the performance of the target model $\mathcal{M}_T$ for the specific task at hand~\cite{TZJRR16,PMGJCS17}. On the other hand, a \emph{reconnaissance adversary} seeks to develop a surrogate model $\mathcal{M}_S$ that approximates $\mathcal{M}_T$ across all possible inputs. Achieving such a high level of fidelity allows the adversary to use the surrogate model for subsequent attacks, such as generating adversarial examples, without needing to interact directly with the target model $\mathcal{M}_T$~\cite{PMGJCS17}.

\textbf{Adversary's Capabilities.}
Firstly, we consider a scenario where the adversary interacts with a target model $\mathcal{M}_T$ via a public API~\cite{TZJRR16,OSF19,HJBGZ21,HWWBSZ21}. The adversary submits a query graph $\mathbf{G}_Q$ and receives responses $\mathbf{R}$, which could be a node embedding matrix ($\mathbf{H}$), a predicted probability matrix ($\boldsymbol{\Theta}$), or a t-SNE projection matrix ($\boldsymbol{\Upsilon}$). These responses represent typical outputs from such APIs, varying in the level of information they provide for tasks like graph visualization, transfer learning, and fine-tuning GNNs.
\par
Additionally, We assume the query graph $\mathbf{G}_Q$, with node features $\mathbf{X}_Q$ and graph structure $\mathbf{A}_Q$, is drawn from the same distribution as the training graph $\mathbf{G}_O$ used to train $\mathcal{M}_T$. This implies that $\mathbf{G}_Q$ and $\mathbf{G}_O$ are sampled from the same dataset, consistent with studies on neural network attacks using public datasets~\cite{JCBKP20,HJBGZ21}, particularly in graph-intensive domains like social networks and molecular structures. As discussed in Section \ref{sec:learning_graph}, this assumption can be adjusted.

\section{Attack Framework}\label{sec:attack_framework}
This section aims to discuss attack scenarios that can be launched by the adversary depending on different levels of knowledge and describe the general framework of GNN model-stealing. We adopt both, the attack structure and taxonomy proposed by Shen and co-authors \cite{shen2021model}.

\subsection{Attack Taxonomy}
{To conduct a model stealing attack, the adversary relies on two essential components: the query graph $\mathbf{G}_Q = (\mathbf{A}_Q, \mathbf{X}_Q, \mathbf{C}_Q)$ and the corresponding response $\mathbf{R}$. The query graph consists of
\begin{itemize}
    \item $\mathbf{A}_Q$: the adjacency matrix, which represents the graph structure,
    \item $\mathbf{X}_Q$: the node feature matrix, detailing the attributes of each node,
    \item $\mathbf{C}_Q$: the labels or class information associated with nodes or edges.
\end{itemize}
The response $\mathbf{R}$ from the target model can be in one of several forms, such as
\begin{itemize}
    \item $\mathbf{H}$: node embeddings, which are vectors representing each node in the graph,
    \item $\boldsymbol{\Theta}$: predicted posterior probabilities, indicating the model's confidence in various outcomes,
    \item $\boldsymbol{\Upsilon}$: a t-SNE projection matrix, which visualizes node embeddings in a 2-dimensional space.
\end{itemize}

The adversary may not have complete structural information about the query graph $\mathbf{G}_Q$, lacking the adjacency matrix $\mathbf{A}_Q$, which leads to six possible attack scenarios grouped into two types. In a Type I attack, the adversary queries the target model using graphs from the same distribution as those used in training. This is common in domains like drug interaction predictions, fraud detection, and recommendation systems, where abundant graph data is available.

Conversely, a Type II attack involves compromising the model without access to the specific graph structure used during training. Here, the adversary must reconstruct or infer missing information, such as in social networks where user connections are private, and only public profile information is accessible~\cite{YLCT19}.

\subsection{Learning the Missing Graph Structure}
\label{sec:learning_graph}

In Type I attacks, the adversary trains a surrogate model $\mathcal{M}_S$ using responses from the target model to the query graph $\mathbf{G}_Q$. In contrast, Type II attacks require inferring the missing graph structure $\mathbf{A}_Q$. Following \cite{shen2021model}, we use the IDGL framework by Chen et al. \cite{CWZ20} to infer $\mathbf{G}_Q$. This involves minimizing a joint loss function that combines task-specific prediction loss with graph regularization. The adversary starts with a k-nearest neighbors ($k$NN) graph based on multi-head weighted cosine similarity and iteratively improves it using IDGL, optimizing the joint loss function without interacting directly with the target model.

\subsection{Training the Surrogate Model}
When the adversary has access to the graph structure $\mathbf{A}_Q$, their goal is to train a surrogate model $\mathcal{M}_S$. This model is trained using the $l$-hop subgraphs of all nodes from the query graph $\mathbf{G}_Q$ as input and the corresponding responses $\mathbf{R}$ from the target model as supervised signals. As noted by \cite{shen2021model}, nodes that are close or connected in $\mathbf{G}_Q$ should also be close in the t-SNE projection matrix $\boldsymbol{\Upsilon}$ or the node embedding matrix $\mathbf{H}$. Thus, the target model's responses can uniformly be treated as embedding vectors. The primary goal is to minimize the loss ($\mathcal{L}_R$) between the surrogate model's output embeddings $\hat{\mathbf{H}}_Q$ and the target model's responses $\mathbf{R}$. It's important to note that the surrogate model's output cannot be directly used for node classification, as it is trained to mimic the target model's responses, not to perform classification. An additional step is required to enable node classification. After training and freezing the surrogate model, an MLP classification head ($\mathcal{O}$) is optimized, using the surrogate model's output ($\hat{\mathbf{H}}_Q$) as input and $\mathbf{C}_Q$ as supervision to minimize prediction error. This two-step process ensures the surrogate model's outputs are effectively utilized for node classification.

\section{Proposed Model Stealing Attack}\label{sec:attack}

This section comprises a description of the training of the surrogate model in model-stealing attacks on GNNs. The proposed method is in line with the framework shown in Section \ref{sec:attack_framework}.

\subsection{Contrastive GNN stealing}

As the number of queries sent by the attacker to the victim's model increases, increase the risk of detection and the cost of the attack. Thus, the efficiency of model stealing is measured by the number of queries sent to the victim model, whose outputs serve as targets for training the surrogate model. However, the adversary faces no constraints on the number of samples sent to the surrogate model. For that reason, we argue that simply matching the outputs of the victim and surrogate model proves inefficient within the context of model stealing. Building upon this observation, we propose a novel approach grounded in contrastive graph learning \cite{Zhu:2020vf, YCSCWS20}. We utilize a contrastive objective function that forces the embedding of corresponding nodes in the surrogate and victim models to converge while simultaneously distinguishing them from the embedding of other nodes within both models.

Starting with the given graph $\mathbf{G}=(\mathbf{X}, \mathbf{A})$, we initiate an augmentation process, resulting in the graph $\mathbf{G'}=(\mathbf{X'}, \mathbf{A'})$. The augmentation is applied to both graph structure and feature matrix resulting in more robust representations that are invariant to various transformations and perturbations. Subsequently, the surrogate model $\mathcal{M}_S$ generates an embedding $\mathbf{H}_S$ based on the augmented graph $\mathbf{G'}$. The dimensional of the embedding $\mathbf{H}_S$ matches that of the response $\mathbf{R}$ obtained from the target model $\mathcal{M}_T$. Utilizing a non-linear projection head $g()$ \cite{pmlr-v119-chen20j,Zhu:2020vf}, we map both the target response $\mathbf{R}$ and the surrogate embedding $\mathbf{H}_S$ into a shared representation space, where a contrastive loss is applied:
\begin{equation}
\mathcal{J}=-\frac{1}{2n} \sum_{i = 1}^{n} [\ell_i({t}, {s})+\ell_i({s}, {t})],
\end{equation}
\begin{equation}    
\ell_i(x, y) = \log  \frac{e^{c(x_i, y_i) / \tau}}{ \sum_{k=1}^{n} e^{c({x}_i, {y}_k) / \tau} + \sum_{k=1, k\neq i}^{n}  e^{c({y}_i, {y}_k) / \tau}}.
\end{equation}
where $t = g(\mathbf{R})$ and $s = g(\mathbf{H}_S)$ represent the matrices in the representation space from the target and surrogate models, respectively, and $x_i$ denotes embedding vector corresponding to node $v_i$ in that space. The function $c$ stands for cosine similarity, and $\tau$ is a temperature parameter.

The embeddings from different models $x_i$ and $y_i$ are a positive pair, drawing closer together through the minimization of $e^{c(x_i, y_i) / \tau}$.

To maintain a distinction between a negative pair of embeddings $x_i$ and $y_i$, $i\neq j$ from target and surrogate models, the term $\sum_{k=1}^{n} e^{c(x_i, y_k) / \tau}$ is employed.

Similarly, embeddings $y_i$ and $y_j$ from the same model form a negative pair for $i\neq j$, which are pushed apart by $\sum_{k=1, k\neq i}^{n} e^{c(y_i, y_k) / \tau}$, which consequently distinguishes embeddings of different nodes in the surrogate embedding space. 

This contrastive objective serves two functions: it enforces the alignment of corresponding node embeddings across surrogate and target models, while simultaneously ensuring their differentiation from embeddings of different nodes within the surrogate model.

Additionally, the augmentations introduce corruption to the original graph within the input of the surrogate model, which generates novel node contexts to contrast with.

\begin{figure}[!t]
\centering
\includegraphics[trim={1.5cm 0cm 0cm 0cm},clip, width=0.95\linewidth]{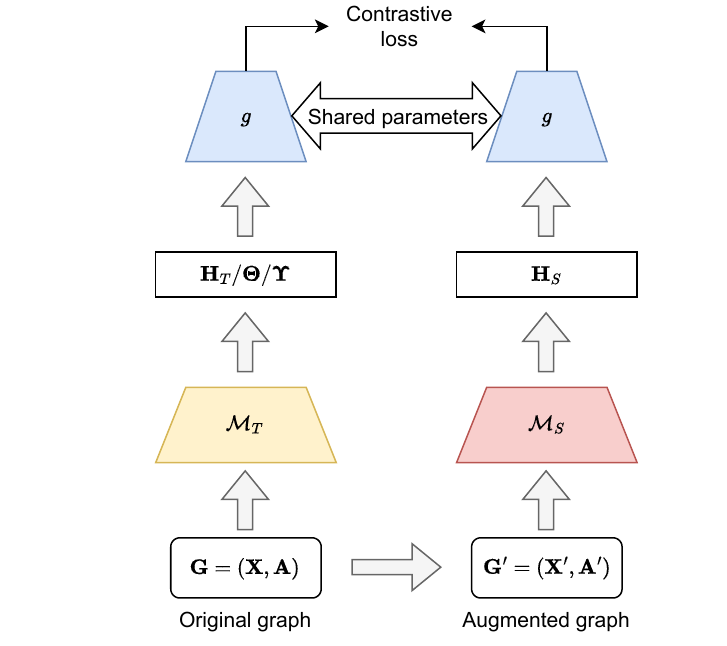} 
\caption{Overview of our model stealing attack against inductive GNNs.}
\vspace{0.7cm}
\label{fig:e2e}
\end{figure}

\subsection{Spectral Graph Augmentations}

One of our key observations is that the existing GNN stealing methods overlook the capability of GNN models to capture essential node information while filtering out irrelevant noise. This results in ineffective utilization of the information received from the victim model. We address it in our approach. In essence, we leverage the basic but essential fact that the victim model produces similar outputs for a given node and its augmented version. Thus, we train the surrogate model to align its predictions with the output of the victim model for both original and augmented nodes. 

Specifically, we consider the graph structure of the GNN training data to develop an optimal augmentation strategy. During surrogate model training, we incorporate spectral graph augmentations initially introduced by \cite{NEURIPS2022_13b45b44}. The augmented graph is constructed by extracting information with varying frequencies from the original graph.

Let $\mathbf{D} = \text{diag}(d_1, \ldots, d_n)$ represent the node degree matrix, where $d_i = \sum_{j \in V} \mathbf{A}_{ij}$ is the degree of node $i \in V$. We define $\mathcal{L} = \mathbf{D} - \mathbf{A}$ as the unnormalized graph Laplacian of $\mathbf{G}$, and $\hat{\mathcal{L}} = I_n - \hat{A} = D^{-\frac{1}{2}}(D - A)D^{-\frac{1}{2}}$ as the symmetric normalized graph Laplacian.
Since $\hat{L}$ is symmetric normalized, its eigen-decomposition is $U\Lambda U^\top$, where $\Lambda = \text{diag}(\lambda_1, \ldots, \lambda_N)$ and $U = [u_1^\top, \ldots, u_N^\top] \in \mathbb{R}^{N \times N}$ are the eigenvalues and eigenvectors of $\hat{L}$, respectively. 
Assuming $0 = \lambda_1 \leq \ldots \leq \lambda_N < 2$ (in which we approximately estimate $\lambda_N \approx 2$ \cite{kipf2017semi}) , we call $\{\lambda_1, \ldots, \lambda_{\lfloor N/2 \rfloor}\}$ the amplitudes of low-frequency components and $\{\lambda_{\lfloor N/2 \rfloor + 1}, \ldots, \lambda_N\}$ the amplitudes of high-frequency components.
The graph spectrum, denoted as $\phi(\lambda)$, is defined by these amplitudes of different frequency components, indicating which parts of the frequency are enhanced or weakened.

For two random augmentations $V_1$ and $V_2$, their graph spectra are $\phi_{V_1}(\lambda)$ and $\phi_{V_2}(\lambda)$. Then, for all $\lambda_m \in [1,2]$ and $\lambda_n \in [0,1]$, $V_1$ and $V_2$ constitute an effective pair of graph augmentations if the following condition is met:
\begin{equation}
Optim_{V_1,V_2}:
|\phi_{V_1}(\lambda_m) - \phi_{V_2}(\lambda_m)| > |\phi_{V_1}(\lambda_n) - \phi_{V_2}(\lambda_n)|\text{.}    
\end{equation}
Such a pair of augmentations is termed an \textit{optimal contrastive pair} as in \citep{NEURIPS2022_13b45b44}. On a high level, ensuring that the difference in high-frequency components is greater than in low-frequency components helps the model to learn more robust and discriminative features.

Following this rule, we transform the adjacency matrix $A$ to a new augmentation $A'$, where $A$ and $A'$ form an optimal contrastive pair $Optim_{A,A'}$ for all  $\lambda_m \in [1,2]$ and $\lambda_n \in [0,1]$. 
The new graph, with its adjacency matrix $A'$, undergoes further augmentations such as feature masking, edge dropping, and edge perturbation. The result serves as the input for training the surrogate model. This form of augmentation improves the learning process of the surrogate model by boosting the effects of contrastive graph learning.

\section{Experimental Evaluation}\label{sec:experiments}
In this section, we conduct a thorough experimental analysis of a model stealing attack on inductive GNNs and show that our approach outperforms the state-of-the-art algorithm by Shen et al.~\cite{shen2021model}. In what follows, we introduce the experimental setup, evaluate Type I and Type II attacks from theft and reconnaissance perspectives, and investigate the impact of different query budgets on attack performance. To our knowledge, ours and Shen et al.~\cite{shen2021model} are the only existing model stealing methods applicable to inductive GNN.

\textbf{Datasets.}
We evaluate performance of our attack on 6 benchmark datasets for evaluating GNN performance~\cite{KW17,HYL17,XHLJ19}: DBLP~\cite{PWZZW16}, Pubmed~\cite{SNBGGE08}, Citeseer Full (referred to as Citeseer)~\cite{GBL98}, Coauthor Physics (referred to as Coauthor)~\cite{SMBG18}, ACM~\cite{WJSWYCY19}, and Amazon Co-purchase Network for Photos (referred to as Amazon)~\cite{MTSH15}. 
We use these datasets to evaluate our attack's efficacy across different graph characteristics such as size, node features, and classes. Each dataset is split into three segments: the first 20\% of randomly selected nodes for training the target model $\mathcal{M}_T$, the next 30\% as the query graph $\mathbf{G}_Q$, and the final 50\% as testing data for both $\mathcal{M}_T$ and $\mathcal{M}_S$. Further, we show that our attacks remain effective even with a reduced number of nodes in the query graph (see Section \ref{sec:query}).
Importantly, in the official implementation of \cite{shen2021model} the testing data is sampled from the whole dataset, creating an overlap between the first two and the third data split. We conduct all experiments, including the state-of-the-art baseline by Shen et al.~\cite{shen2021model} on disjoint data sets.

\textbf{Target and Surrogate Models.}
Following \cite{shen2021model} we use GIN, GAT, and GraphSAGE as our target and surrogate models' architectures for evaluating our attack. 
We outline the model and training details for reproducibility purposes in Appendix \ref{sec:target_model} and \ref{sec:surrogate_model}.

\textbf{Metrics.}
Following ~\cite{JCBKP20} taxonomy, we employ three metrics — accuracy, F1 score and fidelity — to assess our attack's performance. For the theft adversary, we gauge accuracy (correct predictions divided by total predictions) and F1 score (the harmonic mean of precision and recall), a widely-used metric in GNN node classification evaluation~\cite{KW17,HYL17,VCCRLB18}.
As the reconnaissance adversary aims to mimic the target model's behavior, we utilize fidelity (predictions agreed upon by both $\mathcal{M}_S$ and $\mathcal{M}_T$) as our evaluation metric~\cite{JSMA19,JCBKP20}.

\begin{table}[b]
\centering
\caption{The performance of the original classification tasks on all 6 datasets using 3 different GNN structures.}
\vspace{0.5cm}
\scalebox{0.8}{
\begin{tabular}{lcccccc}
\toprule
\multirow{2}{*}{\textbf{Dataset}} 
                 &    \multicolumn{2}{c}{\textbf{GIN}} & 
  \multicolumn{2}{c}{\textbf{GAT}} & 
  \multicolumn{2}{c}{\textbf{SAGE}} \\
 \cmidrule(lr){2-3} \cmidrule(lr){4-5} \cmidrule(lr){6-7}
   &
  \textbf{Accuracy} &
  \textbf{F1} &
  \textbf{Accuracy} &
  \textbf{F1} &
  \textbf{Accuracy} &
  \textbf{F1}\\
\midrule
\textbf{DBLP}  &  0.772 & 0.713 & 0.759 & 0.693 & 0.781 & 0.715 \\
\textbf{Pubmed}  &  0.844 & 0.839 & 0.821 & 0.816 & 0.856 & 0.852 \\
\textbf{Citeseer}  &  0.834 & 0.836 & 0.836 & 0.838 & 0.838 & 0.839\\
\textbf{Coauthor}  &  0.912 & 0.907 & 0.941 & 0.923 & 0.943 & 0.924 \\
\textbf{ACM}  &  0.890 & 0.890 & 0.897 & 0.896 & 0.905 & 0.905 \\
\textbf{Amazon}  &  0.819 & 0.754 & 0.920 & 0.912 & 0.902 & 0.897 \\
\bottomrule
\end{tabular}

}
\label{tab:acc_target_models}
\end{table}

\subsection{Evaluation of Type I Attacks}

\begin{table*}[!t]
\centering
\caption{
The accuracy and fidelity scores of \textbf{Type I} attacks on all of the datasets with \textbf{GIN} as the target model. }
\vspace{0.5cm}

\scalebox{0.8}{
\begin{tabular}{lclccccccccc}
\toprule
\multirow{3}{*}{\textbf{Dataset}} &
  \multirow{3}{*}{\begin{tabular}[x]{@{}c@{}} \textbf{Task} \end{tabular}} 
& & 
  \multicolumn{3}{c}{\textbf{Surrogate GIN}} & 
  \multicolumn{3}{c}{\textbf{Surrogate GAT}} & 
  \multicolumn{3}{c}{\textbf{Surrogate SAGE}} \\
 \cmidrule(lr){4-6} \cmidrule(lr){7-9} \cmidrule(lr){10-12}
 & &
   &
  \textbf{Accuracy} &
  \textbf{Fidelity} &
  \textbf{F1 score} &
  \textbf{Accuracy} &
  \textbf{Fidelity} &
  \textbf{F1 score} &
  \textbf{Accuracy} &
  \textbf{Fidelity} &
  \textbf{F1 score}\\

\midrule
             & &  Shen et al.  &  0.738$\pm$0.001 & 0.805$\pm$0.000 & 0.675$\pm$0.002 & 0.715$\pm$0.003 & 0.782$\pm$0.004 & 0.620$\pm$0.002 & 0.735$\pm$0.001 & 0.799$\pm$0.002 & 0.679$\pm$0.001 \\

            & \multirow{-2}{*}{Prediction} &  \textbf{ours} & \bf 0.774$\pm$0.002 & \bf0.826$\pm$0.002 & \bf0.718$\pm$0.002&\bf0.776$\pm$0.001 &\bf 0.847$\pm$0.002 &\bf 0.717$\pm$0.002 &\bf0.781$\pm$0.001 &\bf 0.834$\pm$0.001 &\bf 0.723$\pm$0.002 \\
                
              \addlinespace

            & &  Shen et al.  & 0.705$\pm$0.007 & 0.751$\pm$0.000 &0.569$\pm$0.004& 0.695$\pm$0.016 & 0.757$\pm$0.020 &0.549$\pm$0.011& 0.704$\pm$0.031 & 0.755$\pm$0.038&\bf 0.588$\pm$0.022\\
            
              & \multirow{-2}{*}{Projection} &  	\textbf{ours} &\bf 0.740$\pm$0.001 &\bf0.783$\pm$0.000 &\bf0.607$\pm$0.004&\bf0.721$\pm$0.004 &\bf0.772$\pm$0.003&\bf0.593$\pm$0.001&\bf0.746$\pm$0.002 &\bf 0.788$\pm$0.001&0.579$\pm$0.004\\

              \addlinespace
            & &  Shen et al.  &  0.705$\pm$0.002 & 0.759$\pm$0.000 &0.626$\pm$0.001& 0.683$\pm$0.006 & 0.723$\pm$0.013 &0.621$\pm$0.001& 0.710$\pm$0.002 & 0.762$\pm$0.001&0.645$\pm$0.004\\
   
\multirow{-6}{*}{\textbf{DBLP}} & \multirow{-2}{*}{Embedding} &  \textbf{ours}  &\bf  0.770$\pm$0.000 &\bf 0.806$\pm$0.004 &\bf 0.804$\pm$0.001&\bf0.774$\pm$0.004 &\bf 0.827$\pm$0.005 &\bf 0.836$\pm$0.002&\bf0.755$\pm$0.009 &\bf 0.796$\pm$0.005&\bf 0.787$\pm$0.002\\
\midrule

                & &  Shen et al.  & 0.846$\pm$0.001 & \textbf{0.912$\pm$0.001} &0.843$\pm$0.004& 0.825$\pm$0.002& \textbf{0.906$\pm$0.001} &0.819$\pm$0.005& 0.846$\pm$0.001&\textbf{0.914$\pm$0.000}&0.840$\pm$0.005\\

              & \multirow{-2}{*}{Prediction} &  \textbf{ours} & \textbf{0.856$\pm$0.001} & 0.892$\pm$0.001 &\textbf{0.847$\pm$0.004}&\textbf{0.845$\pm$0.001} & 0.897$\pm$0.001 &\bf 0.835$\pm$0.004&\textbf{0.859$\pm$0.002} & 0.896$\pm$0.002 &\textbf{0.852$\pm$0.001}\\
                
              \addlinespace

              & &  Shen et al.  & 0.816$\pm$0.026&0.872$\pm$0.029 &0.795$\pm$0.005& 0.797$\pm$0.037&0.864$\pm$0.052 &0.808$\pm$0.005& 0.793$\pm$0.016&0.848$\pm$0.020 &0.833$\pm$0.005\\

              & \multirow{-2}{*}{Projection} &  \textbf{ours} &\textbf{0.848$\pm$0.001} & \textbf{0.883$\pm$0.000} &\textbf{0.841$\pm$0.009}&\textbf{0.840$\pm$0.002} &\textbf{0.896$\pm$0.001} &\bf 0.824$\pm$0.004&\textbf{0.855$\pm$0.002} &\textbf{0.889$\pm$0.002} & \bf0.840$\pm$0.002\\

              \addlinespace
              & &  Shen et al.  &  0.853$\pm$0.002&\textbf{0.910$\pm$0.001} &0.847$\pm$0.002& 0.834$\pm$0.001&\textbf{0.902$\pm$0.005} &0.829$\pm$0.002& 0.846$\pm$0.001&\textbf{0.911$\pm$0.003} &0.839$\pm$0.002\\

\multirow{-6}{*}{\textbf{Pubmed}}  & \multirow{-2}{*}{Embedding} &  \textbf{ours} &\textbf{0.863$\pm$0.002} & 0.888$\pm$0.003 &\textbf{0.857$\pm$0.004}&\textbf{0.853$\pm$0.001} & 0.898$\pm$0.003 &\bf 0.850$\pm$0.009&\textbf{0.862$\pm$0.003} & 0.879$\pm$0.005 & \bf 0.862$\pm$0.005\\

\midrule
               & &  Shen et al.  &  0.773$\pm$0.001 & 0.818$\pm$0.001 &0.777$\pm$0.002& 0.725$\pm$0.007 & 0.766$\pm$0.009 &0.741$\pm$0.001& 0.762$\pm$0.002  & 0.794$\pm$0.003 &0.768$\pm$0.001\\

               & \multirow{-2}{*}{Prediction} &  \textbf{ours} & \bf 0.820$\pm$0.003 & \bf0.843$\pm$0.001 & \bf 0.820$\pm$0.003&\bf0.827$\pm$0.003 &\bf 0.861$\pm$0.003 & \bf 0.827$\pm$0.005&\bf0.814$\pm$0.003 &\bf 0.843$\pm$0.002& \bf 0.814$\pm$0.009\\
                
              \addlinespace

               & &  Shen et al.  &  0.668$\pm$0.050 & 0.688$\pm$0.051 &0.681$\pm$0.001& 0.704$\pm$0.012 & 0.733$\pm$0.015 &0.754$\pm$0.005&\bf 0.664$\pm$0.043  &\bf 0.695$\pm$0.053 &0.599$\pm$0.002\\

               & \multirow{-2}{*}{Projection} &  \textbf{ours} &\bf 0.751$\pm$0.003 & \bf0.763$\pm$0.002 &\bf0.740$\pm$0.001&\bf0.766$\pm$0.005 & \bf0.779$\pm$0.004 &\bf 0.760$\pm$0.009&0.631$\pm$0.006 & 0.648$\pm$0.008 & \bf 0.612$\pm$0.001\\

              \addlinespace
               & &  Shen et al.  &  0.736$\pm$0.005 & 0.772$\pm$0.004 &0.748$\pm$0.004& 0.730$\pm$0.004 & 0.771$\pm$0.007 &0.782$\pm$0.001& 0.753$\pm$0.003  & 0.797$\pm$0.002 &0.745$\pm$0.001\\

\multirow{-6}{*}{\textbf{Citeseer}}  & \multirow{-2}{*}{Embedding} &  \textbf{ours} &  \bf0.810$\pm$0.003 & \bf0.827$\pm$0.001 &\bf0.815$\pm$0.001&\bf0.828$\pm$0.000 &\bf 0.852$\pm$0.000 &\bf 0.837$\pm$0.002&\bf0.813$\pm$0.004 & \bf0.818$\pm$0.011 & \bf 0.810$\pm$0.001\\

\midrule
              & &  Shen et al.  &  0.869$\pm$0.002 & 0.808$\pm$0.000 &0.821$\pm$0.001& 0.759$\pm$0.021 & 0.722$\pm$0.028 &0.671$\pm$0.010& 0.842$\pm$0.012 & 0.801$\pm$0.016 &0.789$\pm$0.002\\

               & \multirow{-2}{*}{Prediction} &  \textbf{ours} &\bf 0.915$\pm$0.003 & \bf0.839$\pm$0.002 &\bf0.877$\pm$0.003&\bf0.855$\pm$0.006 &\bf 0.825$\pm$0.016 &\bf 0.796$\pm$0.003&\bf0.881$\pm$0.009 & \bf0.816$\pm$0.005 &\bf 0.856$\pm$0.001\\
                
              \addlinespace

                & &  Shen et al.  &  0.609$\pm$0.055 & \bf0.633$\pm$0.100 &0.512$\pm$0.012& 0.606$\pm$0.052 & \bf0.600$\pm$0.081 &0.559$\pm$0.001&\bf 0.635$\pm$0.021 &\bf 0.646$\pm$0.011 &0.540$\pm$0.010\\

               & \multirow{-2}{*}{Projection} &  \textbf{ours} & \bf0.635$\pm$0.002 & 0.616$\pm$0.007 &\bf0.673$\pm$0.002&\bf0.619$\pm$0.041 & 0.591$\pm$0.042 &\bf 0.564$\pm$0.009&0.524$\pm$0.015 & 0.538$\pm$0.014 &\bf0.613$\pm$0.002\\
               
              \addlinespace
              & &  Shen et al.  &  0.891$\pm$0.005 & 0.820$\pm$0.000 &0.866$\pm$0.004& 0.777$\pm$0.032 & 0.737$\pm$0.024 &0.781$\pm$0.005& 0.877$\pm$0.004 & 0.809$\pm$0.003 &0.822$\pm$0.001\\

\multirow{-6}{*}{\textbf{Amazon}}  & \multirow{-2}{*}{Embedding} &  \textbf{ours}  &\bf0.919$\pm$0.000 & \bf0.828$\pm$0.005 &\bf0.887$\pm$0.001&\bf0.901$\pm$0.009 &\bf 0.825$\pm$0.010 &\bf 0.834$\pm$0.001&\bf0.893$\pm$0.007 &\bf 0.823$\pm$0.002 & \bf 0.861$\pm$0.004\\

\midrule
                & &  Shen et al.  &  0.913$\pm$0.001 & 0.930$\pm$0.000 &0.888$\pm$0.004& 0.864$\pm$0.009 & 0.888$\pm$0.010 &0.844$\pm$0.001& 0.901$\pm$0.006 & 0.918$\pm$0.007 &0.885$\pm$0.002\\

            & \multirow{-2}{*}{Prediction} &  \textbf{ours} & \bf0.942$\pm$0.002 & \bf0.950$\pm$0.002 &\bf0.926$\pm$0.009&\bf0.921$\pm$0.002 & \bf0.942$\pm$0.003 &\bf 0.876$\pm$0.003&\bf0.942$\pm$0.000 &\bf 0.957$\pm$0.001 &\bf0.923$\pm$0.009\\
                
              \addlinespace

               & &  Shen et al.  &  0.790$\pm$0.047 & 0.797$\pm$0.000&0.760$\pm$0.005& 0.832$\pm$0.018 & 0.854$\pm$0.018 &0.725$\pm$0.001& 0.773$\pm$0.068 & 0.784$\pm$0.070 &0.655$\pm$0.001\\

              & \multirow{-2}{*}{Projection} &  \textbf{ours} & \bf0.920$\pm$0.004 & \bf0.926$\pm$0.004 &\bf0.888$\pm$0.002&\bf0.881$\pm$0.013 & \bf0.896$\pm$0.012 &\bf 0.866$\pm$0.001&\bf0.824$\pm$0.008 &\bf 0.836$\pm$0.008 & \bf 0.675$\pm$0.009\\

              \addlinespace
              & &  Shen et al.  &  0.887$\pm$0.004 & 0.902$\pm$0.000 &0.852$\pm$0.004& 0.878$\pm$0.000 & 0.891$\pm$0.002 &0.869$\pm$0.002& 0.910$\pm$0.005 & 0.922$\pm$0.005 &0.882$\pm$0.004\\

\multirow{-6}{*}{\textbf{Coauthor}}  & \multirow{-2}{*}{Embedding} &  \textbf{ours} & \bf0.943$\pm$0.003 & \bf0.948$\pm$0.005 &\bf0.926$\pm$0.001&\bf0.934$\pm$0.001 & \bf0.944$\pm$0.002 &\bf 0.911$\pm$0.004&\bf0.941$\pm$0.001 & \bf0.941$\pm$0.002& \bf 0.915$\pm$0.005\\

\midrule
              & &  Shen et al.  &  0.733$\pm$0.007&0.759$\pm$0.007 &0.736$\pm$0.004& 0.725$\pm$0.040&0.744$\pm$0.051 &0.741$\pm$0.004& 0.750$\pm$0.008&0.793$\pm$0.007 &0.738$\pm$0.004\\

               & \multirow{-2}{*}{Prediction} &  \textbf{ours}  & \bf 0.895$\pm$0.005 & \bf0.913$\pm$0.006 &\bf0.896$\pm$0.001&\bf0.888$\pm$0.008 & \bf0.934$\pm$0.008 &\bf 0.889$\pm$0.003&\bf0.895$\pm$0.004 &\bf 0.928$\pm$0.004 &\bf 0.906$\pm$0.005\\
                
              \addlinespace

              & &  Shen et al.  & \bf 0.794$\pm$0.009&\bf0.810$\pm$0.016 &0.712$\pm$0.004& \bf0.780$\pm$0.028&0.791$\pm$0.036 &0.737$\pm$0.005& 0.801$\pm$0.039&\bf0.831$\pm$0.042 &0.847$\pm$0.004\\

              & \multirow{-2}{*}{Projection} &  \textbf{ours}  &  0.770$\pm$0.011 & 0.798$\pm$0.008 &\bf0.804$\pm$0.004&0.757$\pm$0.004 &\bf 0.793$\pm$0.003 &\bf 0.738$\pm$0.001&\bf0.816$\pm$0.016 & 0.823$\pm$0.020 & \bf 0.853$\pm$0.002\\

              \addlinespace
               & &  Shen et al.  &  0.811$\pm$0.010&\bf0.836$\pm$0.008 &0.789$\pm$0.001& 0.817$\pm$0.010&0.840$\pm$0.011 &0.800$\pm$0.001& 0.835$\pm$0.003&0.865$\pm$0.003 &0.834$\pm$0.004\\

\multirow{-6}{*}{\textbf{ACM}}  & \multirow{-2}{*}{Embedding} &  \textbf{ours}  & \bf 0.842$\pm$0.007 & 0.822$\pm$0.015 &\bf0.882$\pm$0.003&\bf0.857$\pm$0.009 & \bf0.870$\pm$0.009 &\bf 0.845$\pm$0.001&\bf0.882$\pm$0.009 &\bf 0.894$\pm$0.011& \bf 0.874$\pm$0.003\\
\bottomrule
\end{tabular}
}
\vspace{0.3cm}
\label{tab:acc_fidelity_type1_gin}
\end{table*}
\begin{table*}[!t]
\centering
\caption{
The accuracy and fidelity scores of \textbf{Type II} attacks on all of the datasets with \textbf{GIN} as the target model. }
\vspace{0.5cm}

\scalebox{0.8}{
\begin{tabular}{lclccccccccc}
\toprule
\multirow{3}{*}{\textbf{Dataset}} &
  \multirow{3}{*}{\begin{tabular}[x]{@{}c@{}} \textbf{Task} \end{tabular}} 
& & 
  \multicolumn{3}{c}{\textbf{Surrogate GIN}} & 
  \multicolumn{3}{c}{\textbf{Surrogate GAT}} & 
  \multicolumn{3}{c}{\textbf{Surrogate SAGE}} \\
 \cmidrule(lr){4-6} \cmidrule(lr){7-9} \cmidrule(lr){10-12}
 & &
   &
  \textbf{Accuracy} &
  \textbf{Fidelity} &
  \textbf{F1 score} &
  \textbf{Accuracy} &
  \textbf{Fidelity} &
  \textbf{F1 score} &
  \textbf{Accuracy} &
  \textbf{Fidelity} &
  \textbf{F1 score}\\

\midrule
             & &  Shen et al.   &  0.731$\pm$0.000 & \bf0.822$\pm$0.000 &0.565$\pm$0.004& 0.687$\pm$0.003 & 0.761$\pm$0.004 &0.453$\pm$0.003& 0.732$\pm$0.001 & \bf0.817$\pm$0.000 &0.439$\pm$0.003\\

               & \multirow{-2}{*}{Prediction} &  \textbf{ours}  & \bf0.747$\pm$0.001 & 0.813$\pm$0.001 &\bf \bf0.628$\pm$0.004&\bf0.746$\pm$0.005 &\bf 0.815$\pm$0.003 &\bf0.682$\pm$0.005&\bf0.741$\pm$0.005 & 0.804$\pm$0.005 &\bf0.648$\pm$0.003\\

              \addlinespace

              & &  Shen et al.  &  0.681$\pm$0.001 & 0.752$\pm$0.000 &0.366$\pm$0.001& 0.680$\pm$0.004 & 0.752$\pm$0.005 &\bf0.518$\pm$0.003& 0.677$\pm$0.000 & 0.751$\pm$0.001 &0.362$\pm$0.003\\

              & \multirow{-2}{*}{Projection} &  \textbf{ours} &\bf 0.706$\pm$0.003 &\bf 0.767$\pm$0.003 &\bf \bf0.574$\pm$0.001&\bf0.697$\pm$0.007 & \bf0.753$\pm$0.005 &0.505$\pm$0.001&\bf0.693$\pm$0.006 & \bf0.756$\pm$0.011 &\bf0.579$\pm$0.002\\

              \addlinespace
               & &  Shen et al.   &  0.720$\pm$0.006 &\bf 0.796$\pm$0.000 &0.540$\pm$0.005& 0.682$\pm$0.003 & 0.751$\pm$0.001 &0.540$\pm$0.004&\bf 0.741$\pm$0.002 & \bf0.818$\pm$0.003 &0.605$\pm$0.005\\

\multirow{-6}{*}{\textbf{DBLP}} & \multirow{-2}{*}{Embedding} &  \textbf{ours}  &\bf  0.742$\pm$0.002 & 0.785$\pm$0.002 &\bf \bf0.650$\pm$0.005&\bf0.749$\pm$0.005 &\bf 0.791$\pm$0.002 &\bf0.677$\pm$0.004&0.704$\pm$0.011 & 0.744$\pm$0.009 &\bf0.613$\pm$0.001\\

\midrule
               & &  Shen et al.   &  0.786$\pm$0.001 & 0.831$\pm$0.000 &0.738$\pm$0.001& 0.731$\pm$0.005 & 0.785$\pm$0.003 &0.704$\pm$0.001& 0.785$\pm$0.001 & 0.820$\pm$0.000 &0.691$\pm$0.003\\

              & \multirow{-2}{*}{Prediction} &  \textbf{ours}  & \bf0.853$\pm$0.001 & \bf0.889$\pm$0.001 &\bf0.847$\pm$0.004&\bf0.834$\pm$0.004 &\bf 0.887$\pm$0.005 &\bf0.822$\pm$0.003&\bf0.848$\pm$0.001 & \bf\bf0.884$\pm$0.001 &\bf0.846$\pm$0.004\\

              \addlinespace

              & &  Shen et al.   & 0.814$\pm$0.002 & 0.874$\pm$0.000 &0.716$\pm$0.001& 0.798$\pm$0.001 & 0.871$\pm$0.001 &0.801$\pm$0.001& 0.822$\pm$0.000 &\bf 0.878$\pm$0.000 &0.687$\pm$0.003\\

              & \multirow{-2}{*}{Projection} &  \textbf{ours}  & \bf 0.846$\pm$0.000 &\bf 0.884$\pm$0.002 &\bf0.836$\pm$0.003&\bf0.830$\pm$0.001 & \bf0.883$\pm$0.002 &\bf0.805$\pm$0.009&\bf0.829$\pm$0.004 & 0.866$\pm$0.005 &\bf0.821$\pm$0.002\\

              \addlinespace
               & &  Shen et al.  &  0.831$\pm$0.002 &\bf 0.887$\pm$0.000 &0.774$\pm$0.002& 0.821$\pm$0.002 & \bf0.889$\pm$0.006 &0.819$\pm$0.003& 0.818$\pm$0.000 & 0.864$\pm$0.001 &0.676$\pm$0.001\\

\multirow{-6}{*}{\textbf{Pubmed}} & \multirow{-2}{*}{Embedding} &  \textbf{ours}  &\bf 0.855$\pm$0.001 & 0.879$\pm$0.001 &\bf0.857$\pm$0.001&\bf0.849$\pm$0.002 & 0.883$\pm$0.002 &\bf0.839$\pm$0.004&\bf0.850$\pm$0.002 & \bf0.868$\pm$0.003 &\bf0.842$\pm$0.001\\

\midrule
               & &  Shen et al.   &  0.797$\pm$0.001 & 0.854$\pm$0.000 &0.810$\pm$0.004& 0.764$\pm$0.001 & 0.815$\pm$0.001 &0.820$\pm$0.009& 0.797$\pm$0.001 & 0.853$\pm$0.001&0.798$\pm$0.001\\

              & \multirow{-2}{*}{Prediction} &  \textbf{ours}  & \bf0.813$\pm$0.002 & 0.854$\pm$0.004 &\bf0.822$\pm$0.001&\bf0.840$\pm$0.005 &\bf 0.887$\pm$0.004 &\bf0.836$\pm$0.001&\bf\bf0.813$\pm$0.003 &\bf 0.856$\pm$0.005 &\bf0.813$\pm$0.003\\

              \addlinespace

               & &  Shen et al.  &  0.660$\pm$0.006 & 0.696$\pm$0.000 &0.670$\pm$0.003& 0.709$\pm$0.005 & 0.754$\pm$0.003 &0.667$\pm$0.001& 0.651$\pm$0.010 & 0.680$\pm$0.008&\bf0.675$\pm$0.004\\

               & \multirow{-2}{*}{Projection} &  \textbf{ours}  & \bf 0.727$\pm$0.005 & \bf0.747$\pm$0.002 &\bf0.714$\pm$0.009&\bf0.757$\pm$0.002 &\bf 0.784$\pm$0.001 &\bf0.731$\pm$0.001&\bf0.667$\pm$0.048 & \bf0.691$\pm$0.051 &0.583$\pm$0.005\\

              \addlinespace
              & &  Shen et al.   & 0.780$\pm$0.002 & 0.817$\pm$0.000 &0.791$\pm$0.009& 0.770$\pm$0.002 & 0.796$\pm$0.004 &0.760$\pm$0.001& 0.787$\pm$0.003 &\bf 0.818$\pm$0.001 &0.803$\pm$0.001\\

\multirow{-6}{*}{\textbf{Citeseer}} & \multirow{-2}{*}{Embedding} &  \textbf{ours}  & \bf 0.802$\pm$0.006 &\bf0.825$\pm$0.006 &\bf0.800$\pm$0.002&\bf0.838$\pm$0.004 &\bf 0.851$\pm$0.009 &\bf0.844$\pm$0.001&\bf0.791$\pm$0.005 & 0.800$\pm$0.005 &\bf0.808$\pm$0.009\\

\midrule
              & &  Shen et al.  &  0.881$\pm$0.002 & 0.828$\pm$0.000 &0.833$\pm$0.002& \bf0.876$\pm$0.002 & \bf0.839$\pm$0.005 &0.575$\pm$0.001& 0.642$\pm$0.006 & 0.586$\pm$0.007&0.680$\pm$0.009\\

               & \multirow{-2}{*}{Prediction} &  \textbf{ours}  & \bf 0.911$\pm$0.005 & \bf0.833$\pm$0.002 &\bf0.900$\pm$0.002&0.874$\pm$0.024 & 0.813$\pm$0.034 &\bf0.882$\pm$0.003&\bf0.829$\pm$0.008 & \bf0.775$\pm$0.014 &\bf0.774$\pm$0.001\\

              \addlinespace

               & &  Shen et al.  &  0.532$\pm$0.004 & 0.479$\pm$0.000 &0.654$\pm$0.009& \bf0.480$\pm$0.116 &\bf 0.455$\pm$0.122 &0.301$\pm$0.005&\bf 0.500$\pm$0.009 &\bf 0.461$\pm$0.005&\bf0.688$\pm$0.001\\

               & \multirow{-2}{*}{Projection} &  \textbf{ours}  &\bf 0.661$\pm$0.028 &\bf 0.602$\pm$0.029 &\bf0.673$\pm$0.001&0.462$\pm$0.028 & 0.418$\pm$0.025 &\bf0.492$\pm$0.002&0.473$\pm$0.048 & 0.424$\pm$0.047 &0.491$\pm$0.004\\

              \addlinespace
              & &  Shen et al.   &  0.744$\pm$0.005 & 0.683$\pm$0.000 &0.847$\pm$0.001& 0.863$\pm$0.002 & 0.842$\pm$0.001 &0.619$\pm$0.004& 0.525$\pm$0.006 & 0.473$\pm$0.004 &0.725$\pm$0.002\\

\multirow{-6}{*}{\textbf{Amazon}}  & \multirow{-2}{*}{Embedding} &  \textbf{ours}  & \bf 0.914$\pm$0.003 &\bf 0.838$\pm$0.006 &\bf0.883$\pm$0.009&\bf0.901$\pm$0.004 &\bf 0.851$\pm$0.006 &\bf0.880$\pm$0.005&\bf0.893$\pm$0.008 & \bf0.811$\pm$0.003 &\bf0.835$\pm$0.009\\

\midrule
               
            & &  Shen et al.  &  0.942$\pm$0.000 &\bf 0.958$\pm$0.000 &0.910$\pm$0.005& \bf0.889$\pm$0.008 &\bf 0.921$\pm$0.008 &0.910$\pm$0.009& 0.944$\pm$0.000 &\bf 0.961$\pm$0.000&0.931$\pm$0.005\\

              & \multirow{-2}{*}{Prediction} &  \textbf{ours}  & \bf 0.946$\pm$0.002 & 0.952$\pm$0.001 &\bf0.920$\pm$0.001&0.888$\pm$0.002 & 0.917$\pm$0.002 &0.910$\pm$0.001&0.944$\pm$0.000 & 0.953$\pm$0.001 &\bf0.933$\pm$0.003\\

              \addlinespace

               & &  Shen et al.   &  0.848$\pm$0.002 & 0.856$\pm$0.000 &0.876$\pm$0.005& \bf0.850$\pm$0.018 &\bf 0.866$\pm$0.018 &\bf0.861$\pm$0.001& 0.813$\pm$0.001 & 0.824$\pm$0.001 &0.822$\pm$0.002\\

              & \multirow{-2}{*}{Projection} &  \textbf{ours}  & \bf0.933$\pm$0.001 & \bf0.940$\pm$0.000 &\bf0.911$\pm$0.009&0.734$\pm$0.008 & 0.750$\pm$0.009 &0.850$\pm$0.002&\bf0.873$\pm$0.018 & \bf0.884$\pm$0.015 &\bf0.890$\pm$0.001\\

              \addlinespace
              & &  Shen et al.   &\bf 0.946$\pm$0.001 &\bf 0.959$\pm$0.000 &0.916$\pm$0.004& \bf0.923$\pm$0.001 &\bf 0.944$\pm$0.001 &\bf0.930$\pm$0.004& 0.922$\pm$0.002 & 0.936$\pm$0.004 &0.949$\pm$0.001\\

\multirow{-6}{*}{\textbf{Coauthor}}  & \multirow{-2}{*}{Embedding} &  \textbf{ours}  &  0.941$\pm$0.001 & 0.946$\pm$0.001 &0.916$\pm$0.004&0.898$\pm$0.005 & 0.925$\pm$0.005 &0.920$\pm$0.001&\bf0.936$\pm$0.004 & \bf0.941$\pm$0.003 &\bf0.960$\pm$0.003\\

\midrule
             & &  Shen et al.   &  0.905$\pm$0.000 &\bf 0.945$\pm$0.000 &0.901$\pm$0.009& 0.892$\pm$0.000 &\bf 0.959$\pm$0.000 &0.876$\pm$0.002& \bf0.897$\pm$0.000 &\bf 0.943$\pm$0.000&0.900$\pm$0.001\\

               & \multirow{-2}{*}{Prediction} &  \textbf{ours}  & \bf 0.907$\pm$0.005 & 0.913$\pm$0.003 &\bf0.904$\pm$0.002&\bf\bf0.900$\pm$0.002 & 0.926$\pm$0.002 &\bf0.894$\pm$0.001& 0.892$\pm$0.011 & 0.911$\pm$0.016 &\bf0.910$\pm$0.002\\

              \addlinespace

              & &  Shen et al.   &  0.835$\pm$0.000 & 0.858$\pm$0.000 &0.884$\pm$0.002& 0.872$\pm$0.000 & 0.920$\pm$0.000 &0.859$\pm$0.003& 0.876$\pm$0.000 & 0.904$\pm$0.000&0.879$\pm$0.001\\

              & \multirow{-2}{*}{Prediction} &  \textbf{ours}  & \bf 0.907$\pm$0.005 & \bf 0.913$\pm$0.003 &\bf0.904$\pm$0.002&\bf\bf0.900$\pm$0.002 & \bf0.926$\pm$0.002 &\bf0.894$\pm$0.001&\bf0.892$\pm$0.011 & \bf0.911$\pm$0.016 &\bf0.910$\pm$0.002\\

              \addlinespace
               & &  Shen et al.   & \bf 0.906$\pm$0.000 & \bf0.935$\pm$0.000 &\bf0.910$\pm$0.005& 0.830$\pm$0.000 & 0.871$\pm$0.000 &0.875$\pm$0.002& 0.861$\pm$0.000 & 0.882$\pm$0.000 &0.876$\pm$0.004\\

\multirow{-6}{*}{\textbf{ACM}} & \multirow{-2}{*}{Embedding} &  \textbf{ours}  &  0.885$\pm$0.007 & 0.902$\pm$0.008 &0.880$\pm$0.001&\bf0.882$\pm$0.009 & \bf0.914$\pm$0.008 &\bf0.879$\pm$0.002&\bf0.868$\pm$0.006 &\bf 0.887$\pm$0.001 &\bf0.884$\pm$0.001\\
\bottomrule
\end{tabular}
}
\label{tab:acc_fidelity_type1i_gin}
\end{table*}

The overview of the accuracy and F1 score of the original node classification tasks for the target models are shown in \autoref{tab:acc_target_models}. All analyzed GNN models perform well across all datasets, underscoring the effectiveness of jointly considering node features and graph structure for classification. Subsequently, the accuracy and fidelity results for Type I attacks are presented in Table \ref{tab:acc_fidelity_type1_gin}. We focus on the results when the adversary targets the GIN model with the stealing attack. Similar performance patterns with GAT and Sage as target models are detailed in \autoref{sec:type1_performance_appendix} due to space constraints.

\textbf{Accuracy and F1 score.}
In our analysis of the target GIN model, a noteworthy surge in accuracy and F1 score is observed across all six datasets when utilizing various surrogate models for the prediction task. Particularly remarkable is the ACM dataset, where accuracy and F1 register a substantial increase of approximately 15 \%pt. across all cases. This trend extends to the projection task, showcasing enhanced accuracy for most datasets, including DBLP, Pubmed, Citeseer, and Coauthor. Furthermore, our method exhibits a significant improvement in embedding accuracy across all datasets and surrogate models.
Remarkably, our approach outperforms the baseline method for all tasks, datasets, and surrogate models, with the most pronounced benefits evident in the DBLP dataset.

\textbf{Fidelity.}
Examining the fidelity metrics for the target GIN model, we observe a consistent boost in prediction fidelity across all datasets and surrogate models, except for Pubmed, where our results closely align with those produced by the baseline. The most substantial increase in fidelity for the prediction task is witnessed in the ACM dataset, reaching nearly 20 \%pt. with the GAT surrogate model. Additionally, there is an observable enhancement in projection fidelity for the majority of dataset and surrogate model configurations. In the embedding task, our method achieves increased fidelity across all datasets, except for the Pubmed dataset.
Notably, for the DBLP and Coauthor datasets, our method consistently outperforms the baseline in terms of fidelity for all tasks, datasets, and surrogate models.

\textbf{Stability.}
We conduct three runs for each combination, employing different graph partition seeds to evaluate how accuracy and fidelity values deviate from the average (indicated by standard deviation). The consistently low standard deviation values, as evident in \autoref{tab:acc_fidelity_type1_gin}, signify minimal fluctuation. This observation implies that the adversary can reliably steal from target models with statistically stable accuracy and fidelity.

\subsection{Evaluation of Type II Attacks}
Recall that for Type II attacks, the adversary must first construct an adjacency matrix $\mathbf{A}_Q$ for the query graph $\mathbf{G}_Q$ before querying the target model $\mathcal{M}_T$ and executing the model stealing attack. The outcomes of Type II attacks are outlined in \autoref{tab:acc_fidelity_type1i_gin}. We present results for GIN as the target model; however, similar patterns for targeting GAT and GraphSAGE can be found in \autoref{sec:type2_performance_appendix}.

\textbf{Accuracy and F1 score.}
Under the GIN target model, our approach consistently enhances prediction accuracy and F1 score across the majority of datasets and surrogate models. Noteworthy, the accuracy observed with the SAGE surrogate model on the Amazon dataset increases by 18.7 \%pt. This trend persists in the context of projection accuracy, where our method outperforms competing configurations. Similarly, improvements in embedding accuracy and F1 score are evident, particularly on the Amazon dataset with the SAGE surrogate model, showcasing a remarkable 35.8 \%pt. accuracy lead over the baseline.

\textbf{Fidelity.}
In terms of accuracy fidelity, our method either surpasses or closely aligns with the baseline across the majority of configurations, with the most substantial enhancement observed on the Pubmed dataset. This trend extends to projection fidelity, where our approach exhibits notable advantage, particularly pronounced on the Pubmed dataset. In the realm of embedding fidelity, our method consistently matches or outperforms the baseline. Notably, with the SAGE surrogate model on the Amazon dataset, we achieve an increase in embedding fidelity from 46.3\% to 81.1\%.

\textbf{Stability.}
Table \ref{tab:acc_fidelity_type1i_gin} contains consistently low standard deviation values, highlighting the adversary's ability to steal target models with stable accuracy and fidelity in Type II attacks.

\subsection{Query Budget}\label{sec:query}

We explore the efficacy of our attack under varying query budgets, represented by different sizes of the query graph $\mathbf{G}_Q$. We compare the query efficiency of our approach with that of \cite{shen2021model}, focusing on the Citeseer dataset due to space constraints, though similar trends are observed in other datasets. Accuracy and fidelity outcomes for both Type I and Type II attacks on the embedding task (see Figure \ref{fig:query_embedding}) are presented for three surrogate models: GIN, GAT, and SAGE. Plots for prediction and projection tasks are provided in Appendix \ref{sec:appendix_query}. We also compare our results to the accuracy and fidelity scores achieved by the best-performing surrogate model trained on 30\% of dataset nodes using the method of \cite{shen2021model}. Our method achieves the same or higher level of accuracy and fidelity for prediction and embedding responses using only 50\% of the queries required in the baseline method. For the projection task, we can make a similar observation for the GAT surrogate model. Learning from the t-SNE projection requires using a number of queries closer to 30\% to match the performance best surrogate model of the baseline when GIN or SAGE is selected as the surrogate model.

\begin{figure}[h!]
\centering
\includegraphics[trim={2.5cm 2cm 3cm 2.3cm},clip,width=0.9\linewidth]{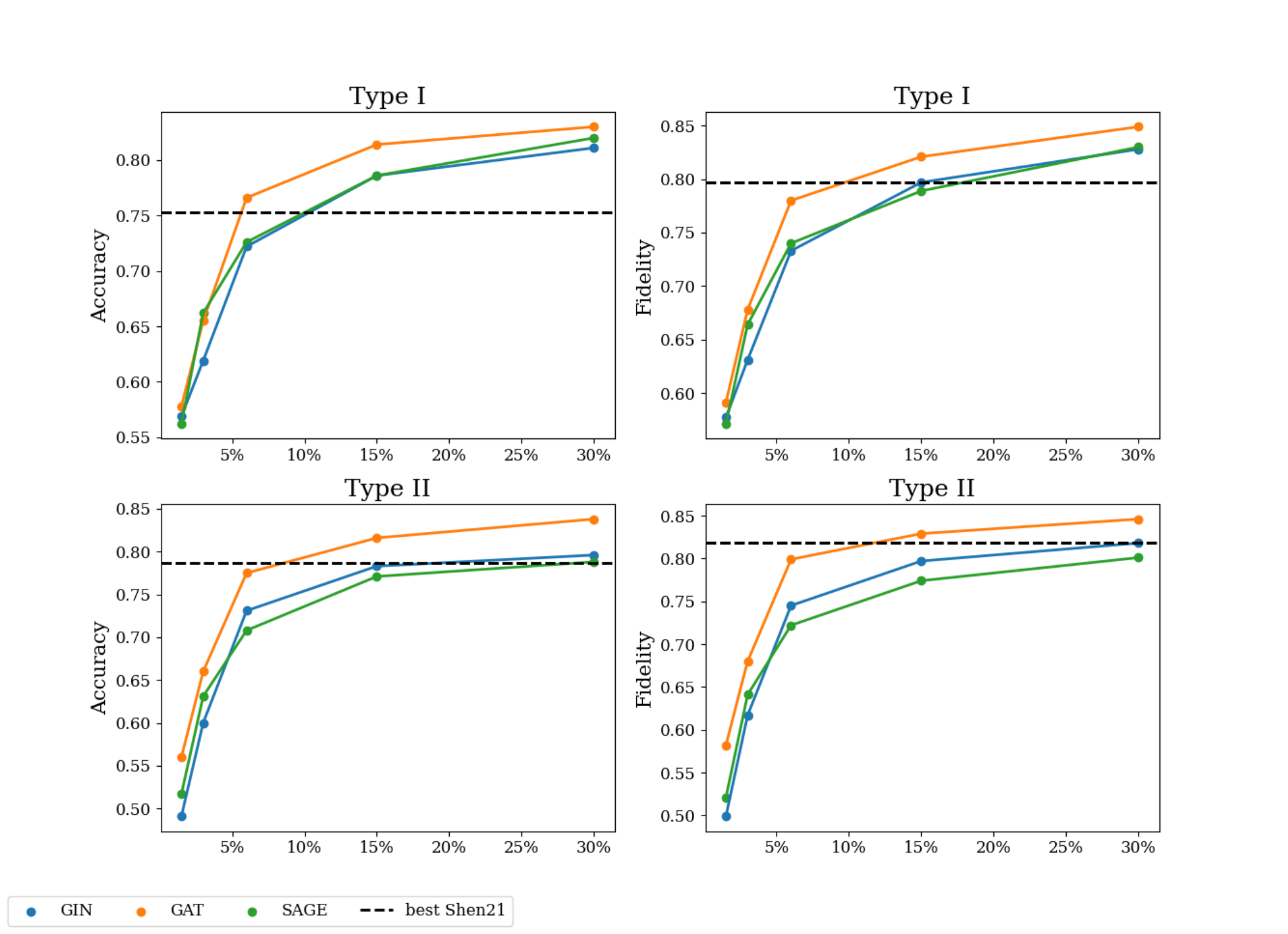} 
\includegraphics[trim={0 0cm 20cm 23cm},clip,width=0.85\linewidth]{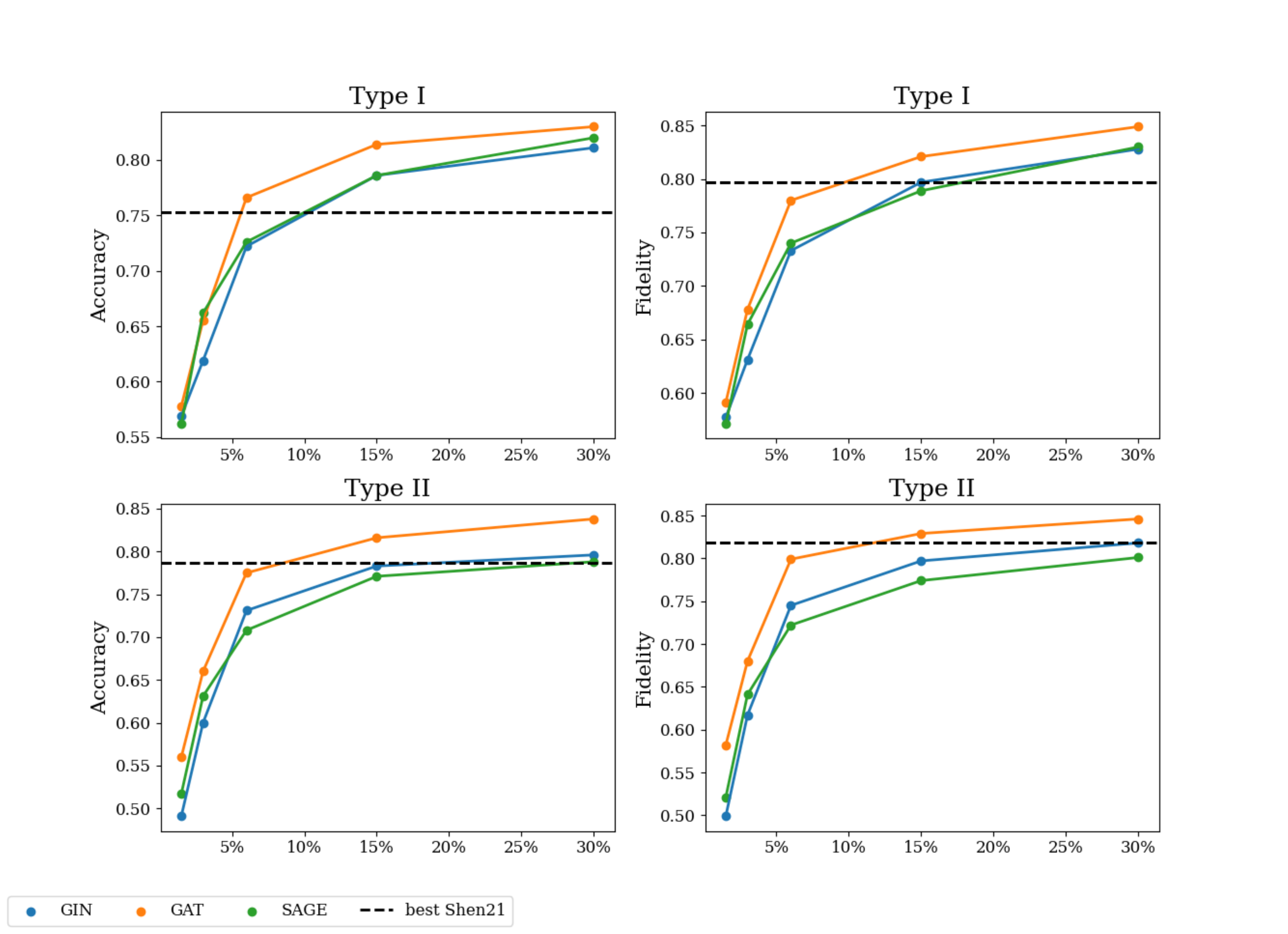} 
\caption{The average accuracies and fidelities on the Citeseer dataset and GIN target with embedding response using different percentages of nodes in the stealing process. We compare with the highest accuracy and fidelity achieved by  \cite{shen2021model} best-performing surrogate model using 30\% of nodes. \vspace{0.4cm}}
\label{fig:query_embedding}
\end{figure}
\section{Discussion}

\textbf{Limitations.}
Firstly, the proposed model stealing attack is limited to scenarios where the target model provides node-level results. However, our method can be extended to graph-level and link prediction tasks by adjusting the augmentations (e.g., using node instead of link augmentations for link prediction). 
Secondly, the learning process of the missing graph structure was based on the approach presented in the paper \cite{shen2021model}. Developing a novel technique for reconstructing graph structure, specifically designed for training the surrogate model with contrastive loss, constitutes a direction for further research.

\textbf{Defense.}
Several active countermeasures against model stealing attacks have been proposed in the literature~\cite{dubinski2024bucks, dziedzic2022increasing, TZJRR16}. The applied formulation of the defense mechanism follows the approach by Shen and co-authors \cite{shen2021model}. To address diverse query responses, we inject random Gaussian noise into node embeddings and t-SNE projections produced by target models, employing GIN as the surrogate model to assess the countermeasure's efficacy. The surrogate model's accuracy, serving as a metric for attack performance within the defined threat model, is evaluated on the ACM dataset (refer to Figure \ref{fig:defense_embedding} for the embedding response and Appendix \ref{sec:appendix_defense} for prediction and projection responses). Our findings indicate a marginal impact of random Gaussian noise on the surrogate model's accuracy, similar to the results obtained by \cite{shen2021model}. Even with a high standard deviation of the noise added, the model stealing attack still results in a useful surrogate model. Moreover, injecting random noise into the target's model responses results in a lowered utility of the model for legitimate users of the model API, as outlined in \cite{pmlr-v162-dziedzic22a}. Developing robust active defense mechanisms continues to be a focus for future research. However, we note the existence of several methods for protecting the Intellectual Property after the GNN model has already been stolen~\cite{watermark1, watermark2, watermark3, watermark4, watermark5, watermark6}.

\begin{figure}[!h]
\centering
\includegraphics[trim={0.2cm -0.cm 1.cm 0cm},clip,width=0.95\linewidth]{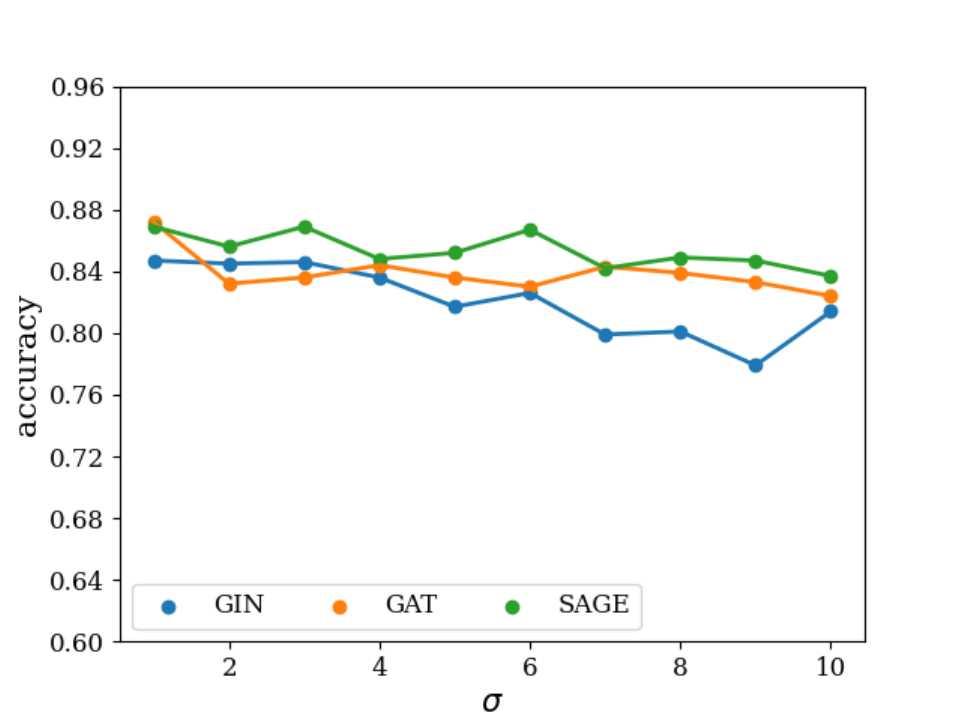} 
\caption{Accuracy scores after adding Gaussian noise with a standard deviation $\sigma$ to embedding response from target model GIN on the ACM dataset.}
\vspace{0.7cm}
\label{fig:defense_embedding}
\end{figure}

\section{Conclusions}

In this work, we identify a new unsupervised GNN model-stealing attack. 
Only a single work~\cite{shen2021model} targeting inductive GNNs existed previously. 
Our proposed approach combines a contrastive learning objective with graph-specific augmentations, improving the efficient utilization of information derived from target model outputs. Specifically, we applied spectral graph augmentations to create varied graph perspectives for the surrogate model within a contrastive learning framework at the node level. Thus, the considered method seamlessly aligns with the overarching framework proposed by \cite{shen2021model}, serving as an enhanced approach for training surrogate models. This adaptation significantly enhances the efficiency and overall performance of the model-stealing process. 
The conducted comparative analysis against the state of the art showed that the proposed stealing technique demonstrates superior performance while requiring fewer queries of the target model. %

\clearpage
\bibliography{main.bib}

\onecolumn

\maketitle

\appendix

\section{Target Model Architectures ($\mathcal{M}_T$)}\label{sec:target_model}
In our evaluation, we employ GIN, GAT, and GraphSAGE as target models. For reproducibility, we provide concise details:

\begin{itemize}
\item \textbf{GIN:}
A 3-layer GIN model with a fixed neighborhood sample size of 10 at each layer. The first hidden layers have 128 hidden units, and the final layer is used for classification.

\item \textbf{GAT:}
A 3-layer GAT model with a fixed neighborhood sample size of 10 at each layer. The first layer consists of 4 attention heads with a hidden unit size of 128. The second layer comprises 4 attention heads with the hidden unit size being the chosen embedding size. The final layer follows the original design for classification~\cite{VCCRLB18}.

\item \textbf{GraphSAGE:}
A 2-layer GraphSAGE following Hamilton et al.~\cite{HYL17}, with neighborhood sample sizes of 25 and 10, respectively. The first hidden layer has a hidden unit size of 128, and the second layer has the hidden unit size set to the chosen embedding size. Each layer employs a GCN aggregator and a 0.5 dropout rate to prevent overfitting. Classification is done using a linear transformation layer.

\end{itemize}

All models utilize cross-entropy as the loss function, ReLU as the activation function between layers, and Adam as the optimizer with an initial learning rate of 0.001. Training spans 200 epochs, and the best models are selected based on the highest validation accuracy. These designs adhere to the specifications outlined in the respective papers.

\section{Surrogate Model Architecture ($\mathcal{M}_S$)}\label{sec:surrogate_model}
For evaluation, we adopt customized GraphSAGE, GAT, and GIN models as our surrogate models. Key details are summarized below:

\begin{itemize}
\item \textbf{GIN:}
A 2-layer GIN model with neighborhood sample sizes of 10 and 50. The hidden unit size for the first layer is 128, and for the second layer, it matches the size of the query response.

\item \textbf{GAT:}
A 2-layer GAT model with neighborhood sample sizes of 10 and 50. Both layers consist of 4 attention heads, following the same hidden unit size as mentioned for GIN.

\item \textbf{GraphSAGE:}
A 2-layer GraphSAGE with neighborhood sample sizes of 10 and 50, following the same hidden unit size as mentioned for GIN.
\end{itemize}

The classification head is a 2-layer MLP with a hidden unit size of 100, taking the output from the first component as input and employing cross-entropy as its loss function. Both components use the Adam optimizer with a learning rate of 0.001. We train the surrogate model and the classification head for 200 and 300 epochs, respectively.

In the augmentation phase we use feature masking with probability $p_1=0.5$, edge dropping with probability $p_2=0.1$, and for the spectral augmentations we use $\epsilon=0.01$, $\eta=1$, $\Omega=20$, and $Iter=3$ following \cite{NEURIPS2022_13b45b44}. In the contrastive learning, we use one layer projection $g()$ with output size 128 and ELU activation function, and loss $\mathcal{J}$ with $\tau=1.0$.

\clearpage
\section{Additional result - Type I attack}\label{sec:type1_performance_appendix}

\begin{table*}[!h]
\centering
\caption{
The accuracy and fidelity scores of \textbf{Type I} attacks on all of the datasets with \textbf{GAT} as the target model. }
\vspace{0.5cm}

\scalebox{0.8}{
\begin{tabular}{lclccccccccc}
\toprule
\multirow{3}{*}{\textbf{Dataset}} &
  \multirow{3}{*}{\begin{tabular}[x]{@{}c@{}} \textbf{Task} \end{tabular}} 
& & 
  \multicolumn{3}{c}{\textbf{Surrogate GIN}} & 
  \multicolumn{3}{c}{\textbf{Surrogate GAT}} & 
  \multicolumn{3}{c}{\textbf{Surrogate SAGE}} \\
 \cmidrule(lr){4-6} \cmidrule(lr){7-9} \cmidrule(lr){10-12}
 & &
   &
  \textbf{Accuracy} &
  \textbf{Fidelity} &
  \textbf{F1 score} &
  \textbf{Accuracy} &
  \textbf{Fidelity} &
  \textbf{F1 score} &
  \textbf{Accuracy} &
  \textbf{Fidelity} &
  \textbf{F1 score}\\

\midrule
             & &  Shen et al.   &  0.764$\pm$0.000 & \bf 0.852$\pm$0.000 &0.695$\pm$0.001& \bf 0.772$\pm$0.001 & \bf 0.889$\pm$0.001 &0.688$\pm$0.005& 0.768$\pm$0.001 & \bf  0.855$\pm$0.002&\bf0.698$\pm$0.005\\

               & \multirow{-2}{*}{Prediction} &  \textbf{ours}  &  \bf 0.772$\pm$0.001 & 0.804$\pm$0.001 &\bf0.706$\pm$0.009&0.762$\pm$0.003 & 0.825$\pm$0.004 &\bf0.692$\pm$0.001&\bf 0.778$\pm$0.002 & 0.813$\pm$0.002 &0.691$\pm$0.003\\

              \addlinespace

              & &  Shen et al.  & 0.697$\pm$0.000 & 0.618$\pm$0.000 &0.527$\pm$0.004& 0.720$\pm$0.008 & \bf 0.790$\pm$0.010 &0.604$\pm$0.003& 0.686$\pm$0.033 & 0.742$\pm$0.038&0.363$\pm$0.041\\

              & \multirow{-2}{*}{Projection} &  \textbf{ours}  &  \bf 0.735$\pm$0.002 & \bf 0.771$\pm$0.003 &\bf0.609$\pm$0.001&\bf 0.741$\pm$0.001 & 0.789$\pm$0.002 &\bf0.625$\pm$0.001&\bf 0.726$\pm$0.003 & \bf 0.772$\pm$0.006 &\bf0.541$\pm$0.005\\

              \addlinespace
               & &  Shen et al.   &  0.759$\pm$0.001 & \bf  0.846$\pm$0.000 &0.673$\pm$0.009& \bf 0.776$\pm$0.001 & \bf 0.872$\pm$0.010 &\bf0.706$\pm$0.001& \bf 0.761$\pm$0.005 & \bf  0.863$\pm$0.008 &0.651$\pm$0.005\\

\multirow{-6}{*}{\textbf{DBLP}} & \multirow{-2}{*}{Embedding} &  \textbf{ours}  &  \bf 0.764$\pm$0.002 & 0.793$\pm$0.004 &\bf0.683$\pm$0.009&0.775$\pm$0.002 & 0.816$\pm$0.001 &0.699$\pm$0.002&0.759$\pm$0.004 & 0.790$\pm$0.003 &\bf0.696$\pm$0.001\\

\midrule
                  & &  Shen et al.   &  0.840$\pm$0.001 &  \bf 0.912$\pm$0.000 &0.836$\pm$0.002& 0.824$\pm$0.001 &  \bf 0.931$\pm$0.002 &0.819$\pm$0.003& 0.838$\pm$0.000 & \bf  0.917$\pm$0.000&0.832$\pm$0.003\\

              & \multirow{-2}{*}{Prediction} &  \textbf{ours}  &  \bf 0.852$\pm$0.001 & 0.874$\pm$0.001 &\bf\bf0.847$\pm$0.001& \bf 0.834$\pm$0.001 & 0.886$\pm$0.002 &\bf0.836$\pm$0.004& \bf 0.853$\pm$0.001 & 0.879$\pm$0.001 &\bf0.848$\pm$0.003\\

              \addlinespace

              & &  Shen et al.   & 0.819$\pm$0.008 &  \bf 0.874$\pm$0.000 &0.722$\pm$0.004& 0.812$\pm$0.005 &  \bf 0.903$\pm$0.008 &0.799$\pm$0.005& 0.800$\pm$0.023 & \bf  0.856$\pm$0.024&0.809$\pm$0.001\\

              & \multirow{-2}{*}{Projection} &  \textbf{ours}  &  \bf  0.842$\pm$0.002 & 0.865$\pm$0.002 &\bf0.833$\pm$0.001& \bf 0.827$\pm$0.002 & 0.880$\pm$0.002 &\bf0.815$\pm$0.001& \bf 0.841$\pm$0.001 & 0.867$\pm$0.000 &\bf0.827$\pm$0.003\\

              \addlinespace
               & &  Shen et al.  &  0.843$\pm$0.000 & \bf  0.915$\pm$0.000 &0.839$\pm$0.001& 0.827$\pm$0.002 &  \bf 0.919$\pm$0.002 &0.822$\pm$0.005& 0.838$\pm$0.001 &  \bf 0.930$\pm$0.001  &0.833$\pm$0.009\\

\multirow{-6}{*}{\textbf{Pubmed}} & \multirow{-2}{*}{Embedding} &  \textbf{ours}  &  \bf  0.854$\pm$0.002 & 0.871$\pm$0.000 &\bf0.852$\pm$0.009& \bf 0.836$\pm$0.002 & 0.880$\pm$0.003 &\bf0.831$\pm$0.003& \bf 0.854$\pm$0.001 & 0.859$\pm$0.001 &\bf0.857$\pm$0.004\\

\midrule
               & &  Shen et al.   &   \bf 0.821$\pm$0.002 & \bf  0.892$\pm$0.000 &\bf0.824$\pm$0.005&   0.827$\pm$0.003 &  \bf 0.910$\pm$0.001 &0.827$\pm$0.001&  \bf 0.825$\pm$0.003 & 0.884$\pm$0.002&\bf0.830$\pm$0.003\\

               & \multirow{-2}{*}{Prediction} &  \textbf{ours}  &  0.820$\pm$0.003 & 0.849$\pm$0.003 &0.821$\pm$0.002&\bf0.835$\pm$0.003 & 0.863$\pm$0.005 &\bf0.842$\pm$0.001&0.820$\pm$0.001 &  \bf 0.855$\pm$0.002 &0.822$\pm$0.005\\

              \addlinespace

               & &  Shen et al.  & 0.626$\pm$0.028 & 0.645$\pm$0.000 &0.657$\pm$0.001& 0.744$\pm$0.005 &  \bf 0.780$\pm$0.001 &\bf0.753$\pm$0.001& 0.649$\pm$0.019 & 0.672$\pm$0.017&0.694$\pm$0.001\\

               & \multirow{-2}{*}{Projection} &  \textbf{ours}  &  \bf  0.731$\pm$0.005 &  \bf 0.749$\pm$0.005 &\bf0.724$\pm$0.001& \bf 0.753$\pm$0.004 & 0.777$\pm$0.005 &0.747$\pm$0.001& \bf 0.686$\pm$0.007 &  \bf 0.689$\pm$0.008 &\bf0.696$\pm$0.004\\

              \addlinespace
               & &  Shen et al.   &   \bf 0.825$\pm$0.003 &  \bf 0.903$\pm$0.000 &0.821$\pm$0.009&  \bf 0.843$\pm$0.002 &  \bf 0.912$\pm$0.001 &0.845$\pm$0.002& 0.816$\pm$0.003 &  \bf 0.882$\pm$0.003 &0.813$\pm$0.004\\

\multirow{-6}{*}{\textbf{Citeseer}} & \multirow{-2}{*}{Embedding} &  \textbf{ours}  & 0.819$\pm$0.003 & 0.842$\pm$0.003 &\bf0.823$\pm$0.005&0.842$\pm$0.006 & 0.866$\pm$0.005 &\bf0.848$\pm$0.005& \bf 0.826$\pm$0.004 & 0.839$\pm$0.006 &\bf0.824$\pm$0.001\\

\midrule
              & &  Shen et al.  &  \bf  0.935$\pm$0.000 &  \bf 0.945$\pm$0.000 &\bf0.923$\pm$0.001& \bf  0.916$\pm$0.002 &  \bf 0.933$\pm$0.004 &\bf0.906$\pm$0.009& \bf  0.916$\pm$0.003 &  \bf 0.936$\pm$0.004&\bf0.897$\pm$0.002\\

              & \multirow{-2}{*}{Prediction} &  \textbf{ours}  &  0.925$\pm$0.002 & 0.929$\pm$0.001 &0.912$\pm$0.003&0.900$\pm$0.012 & 0.917$\pm$0.012 &0.853$\pm$0.009&0.914$\pm$0.001 & 0.921$\pm$0.003 &0.877$\pm$0.009\\

              \addlinespace

               & &  Shen et al.  & 0.671$\pm$0.020 & 0.664$\pm$0.000 &0.552$\pm$0.002& 0.717$\pm$0.026 & 0.726$\pm$0.025 &0.592$\pm$0.012& 0.683$\pm$0.023 & 0.683$\pm$0.028&0.549$\pm$0.003\\

               & \multirow{-2}{*}{Projection} &  \textbf{ours}  &  \bf  0.784$\pm$0.011 &  \bf 0.785$\pm$0.011 &\bf0.708$\pm$0.005& \bf 0.756$\pm$0.014 &  \bf 0.764$\pm$0.014 &\bf0.681$\pm$0.001& \bf 0.711$\pm$0.018 &  \bf 0.708$\pm$0.022 &\bf0.572$\pm$0.003\\

              \addlinespace
              & &  Shen et al.   &  \bf 0.917$\pm$0.000 &  \bf 0.928$\pm$0.000 &\bf0.892$\pm$0.009&  \bf 0.918$\pm$0.001 &  \bf 0.940$\pm$0.001 &\bf0.908$\pm$0.002& 0.880$\pm$0.004 & \bf  0.903$\pm$0.004  &0.850$\pm$0.003\\

\multirow{-6}{*}{\textbf{Amazon}}  & \multirow{-2}{*}{Embedding} &  \textbf{ours}  & 0.915$\pm$0.003 & 0.921$\pm$0.001 &0.889$\pm$0.004&0.899$\pm$0.005 & 0.907$\pm$0.008 &0.883$\pm$0.003& \bf 0.890$\pm$0.006 & 0.895$\pm$0.003 &\bf0.851$\pm$0.002\\

\midrule
               & &  Shen et al.  &  0.940$\pm$ 0.000 &  0.954$\pm$0.000 &0.919$\pm$0.005&  0.926$\pm$ 0.002 & 0.927$\pm$ 0.002 &\bf0.908$\pm$0.001&   \bf 0.942$\pm$ 0.000 &   \bf 0.962$\pm$ 0.001 &\bf0.922$\pm$0.009\\

              & \multirow{-2}{*}{Prediction} &  \textbf{ours}  &   \bf 0.946$\pm$0.001 & 0.954$\pm$0.002 &\bf0.929$\pm$0.004& \bf 0.935$\pm$0.005 &  \bf 0.950$\pm$0.005 &0.905$\pm$0.004&0.939$\pm$0.001 & 0.951$\pm$0.001 &0.919$\pm$0.002\\

              \addlinespace

               & &  Shen et al.   &  0.792$\pm$ 0.066 &  0.801$\pm$ 0.069 &0.593$\pm$0.001&  0.864$\pm$ 0.005 & 0.874$\pm$ 0.005 &0.823$\pm$0.001&  0.727$\pm$ 0.038 &  0.734$\pm$ 0.941 &0.650$\pm$0.003\\

              & \multirow{-2}{*}{Projection} &  \textbf{ours}  &  \bf  0.905$\pm$0.008 & \bf  0.912$\pm$0.010 &\bf\bf0.847$\pm$0.001& \bf 0.898$\pm$0.008 &  \bf 0.904$\pm$0.008 &\bf0.868$\pm$0.001& \bf \bf\bf0.883$\pm$0.003 & \bf  0.894$\pm$0.004 &\bf0.827$\pm$0.004\\

              \addlinespace
              & &  Shen et al.   &  0.932$\pm$ 0.002 &  0.945$\pm$ 0.002 &0.910$\pm$0.005&  0.938$\pm$ 0.002 & 0.927$\pm$ 0.002 &0.916$\pm$0.001&  0.935$\pm$ 0.001 &  \bf  0.953$\pm$ 0.001 &0.912$\pm$0.002\\

\multirow{-6}{*}{\textbf{Coauthor}}  & \multirow{-2}{*}{Embedding} &  \textbf{ours}  &  \bf  0.941$\pm$0.002 &  \bf 0.946$\pm$0.002 &\bf0.922$\pm$0.001& \bf 0.942$\pm$0.002 & \bf  0.949$\pm$0.003 &\bf0.927$\pm$0.009& \bf 0.942$\pm$0.003 & 0.949$\pm$0.002 &\bf0.923$\pm$0.003\\

\midrule
              & &  Shen et al.   &  0.898$\pm$0.004 &  \bf 0.926$\pm$0.000 &0.896$\pm$0.004& 0.889$\pm$0.001 &  \bf 0.939$\pm$0.004 &0.883$\pm$0.002& 0.893$\pm$0.001 &  \bf 0.934$\pm$0.001&0.895$\pm$0.001\\

               & \multirow{-2}{*}{Prediction} &  \textbf{ours}  & \bf  0.901$\pm$0.004 & 0.922$\pm$0.006 &\bf0.898$\pm$0.002& \bf 0.896$\pm$0.001 & 0.935$\pm$0.005 &\bf0.895$\pm$0.002& \bf 0.901$\pm$0.002 & 0.923$\pm$0.004 &\bf0.900$\pm$0.009\\

              \addlinespace

              & &  Shen et al.   &  0.860$\pm$0.010 &  \bf 0.900$\pm$0.000 &0.765$\pm$0.002& 0.885$\pm$0.004 & 0.916$\pm$0.006 &0.881$\pm$0.002& 0.851$\pm$0.029 & 0.883$\pm$0.020&0.873$\pm$0.001\\

              & \multirow{-2}{*}{Projection} &  \textbf{ours}  &   \bf 0.871$\pm$0.006 & 0.892$\pm$0.006 &\bf\bf0.883$\pm$0.003& \bf 0.887$\pm$0.001 & \bf  0.937$\pm$0.002 &\bf0.882$\pm$0.009& \bf 0.870$\pm$0.014 &  \bf 0.894$\pm$0.019 &\bf0.898$\pm$0.005\\

              \addlinespace
              & &  Shen et al.   &  \bf  0.898$\pm$0.003 &  \bf 0.931$\pm$0.000 &\bf0.887$\pm$0.002& 0.893$\pm$0.002 &  \bf 0.937$\pm$0.002 &0.892$\pm$0.005& 0.831$\pm$0.012 & 0.855$\pm$0.025  &0.833$\pm$0.004\\

\multirow{-6}{*}{\textbf{ACM}} & \multirow{-2}{*}{Embedding} &  \textbf{ours}  &  0.870$\pm$0.011 & 0.862$\pm$0.009 &0.877$\pm$0.001& \bf 0.894$\pm$0.005 & 0.928$\pm$0.003 &\bf0.899$\pm$0.004& \bf 0.879$\pm$0.010 & \bf  0.881$\pm$0.010 &\bf0.871$\pm$0.004\\
\bottomrule
\end{tabular}
}
\label{tab:acc_fidelity_type1_gat}
\end{table*}

\clearpage

\begin{table*}[!t]
\centering
\caption{
The accuracy and fidelity scores of \textbf{Type I} attacks on all of the datasets with \textbf{SAGE} as the target model. }
\vspace{0.5cm}

\scalebox{0.8}{
\begin{tabular}{lclccccccccc}
\toprule
\multirow{3}{*}{\textbf{Dataset}} &
  \multirow{3}{*}{\begin{tabular}[x]{@{}c@{}} \textbf{Task} \end{tabular}} 
& & 
  \multicolumn{3}{c}{\textbf{Surrogate GIN}} & 
  \multicolumn{3}{c}{\textbf{Surrogate GAT}} & 
  \multicolumn{3}{c}{\textbf{Surrogate SAGE}} \\
 \cmidrule(lr){4-6} \cmidrule(lr){7-9} \cmidrule(lr){10-12}
 & &
   &
  \textbf{Accuracy} &
  \textbf{Fidelity} &
  \textbf{F1 score} &
  \textbf{Accuracy} &
  \textbf{Fidelity} &
  \textbf{F1 score} &
  \textbf{Accuracy} &
  \textbf{Fidelity} &
  \textbf{F1 score}\\

\midrule
             
                & &  Shen et al.   &  0.776$\pm$0.001 & \bf0.899$\pm$0.000 &0.698$\pm$0.001& 0.779$\pm$0.001 & \bf0.901$\pm$0.003 &0.694$\pm$0.004& 0.777$\pm$0.003 & \bf0.899$\pm$0.004&\bf0.697$\pm$0.001\\

                & \multirow{-2}{*}{Prediction} &  \textbf{ours}  &  \bf0.783$\pm$0.001 & 0.855$\pm$0.001 &\bf\bf0.712$\pm$0.003&\bf0.781$\pm$0.002 & 0.848$\pm$0.001 &\bf0.734$\pm$0.003&\bf0.784$\pm$0.001 & 0.857$\pm$0.003 &0.683$\pm$0.001\\

              \addlinespace

              & &  Shen et al.  &  0.714$\pm$0.027 & 0.811$\pm$0.000 &0.557$\pm$0.002& 0.744$\pm$0.002 & \bf0.839$\pm$0.004 &0.604$\pm$0.005& 0.739$\pm$0.014 &  \bf 0.836$\pm$0.025&0.484$\pm$0.012\\

              & \multirow{-2}{*}{Projection} &  \textbf{ours}  &  \bf0.750$\pm$0.001 & \bf0.827$\pm$0.004 &\bf\bf0.618$\pm$0.001&\bf0.754$\pm$0.004 & 0.829$\pm$0.007 &\bf0.620$\pm$0.003& \bf 0.749$\pm$0.005 & 0.821$\pm$0.004 &\bf0.616$\pm$0.005\\

              \addlinespace
               & &  Shen et al.   &  0.773$\pm$0.001 &  \bf 0.878$\pm$0.000 &0.700$\pm$0.005&  \bf 0.781$\pm$0.002 &  \bf 0.879$\pm$0.001 &\bf0.715$\pm$0.001&  \bf 0.783$\pm$0.000 &  \bf 0.903$\pm$0.002  &0.706$\pm$0.003\\

\multirow{-6}{*}{\textbf{DBLP}} & \multirow{-2}{*}{Embedding} &  \textbf{ours}  &  0.773$\pm$0.002 & 0.840$\pm$0.003 &\bf\bf0.715$\pm$0.009&0.779$\pm$0.002 & 0.842$\pm$0.010 &0.706$\pm$0.001&0.763$\pm$0.003 & 0.815$\pm$0.007 &\bf0.707$\pm$0.001\\

\midrule
               & &  Shen et al.   &  \bf 0.859$\pm$0.001 & \bf  0.966$\pm$0.000 &\bf0.856$\pm$0.002& 0.847$\pm$0.000 &  \bf 0.956$\pm$0.001 &0.845$\pm$0.005& 0.854$\pm$0.000 & \bf  0.964$\pm$0.002&0.852$\pm$0.001\\

              & \multirow{-2}{*}{Pre diction} &  \textbf{ours}  &  0.856$\pm$0.002 & 0.920$\pm$0.005 &0.851$\pm$0.003& \bf 0.848$\pm$0.001 & 0.923$\pm$0.002 &\bf\bf0.849$\pm$0.005& \bf 0.858$\pm$0.001 & 0.926$\pm$0.002 & \bf0.854$\pm$0.005\\

              \addlinespace

              & &  Shen et al.   & 0.821$\pm$0.007 & 0.890$\pm$0.000 &0.757$\pm$0.005& 0.841$\pm$0.001 &  \bf 0.945$\pm$0.009 &0.823$\pm$0.004& \bf  0.844$\pm$0.004 & \bf  0.936$\pm$0.003&0.832$\pm$0.003\\

              & \multirow{-2}{*}{Projection} &  \textbf{ours}  &  \bf  0.849$\pm$0.001 &  \bf 0.915$\pm$0.002 &\bf0.842$\pm$0.003& \bf 0.844$\pm$0.001 & 0.922$\pm$0.003 &\bf0.836$\pm$0.004&0.843$\pm$0.001 & 0.911$\pm$0.000 & \bf 0.825$\pm$0.003\\

              \addlinespace
               & &  Shen et al.  &  \bf  0.863$\pm$0.001 & \bf  0.964$\pm$0.000 &\bf0.860$\pm$0.001& 0.848$\pm$0.001 & \bf  0.958$\pm$0.000 &0.847$\pm$0.002& 0.853$\pm$0.002 &  \bf 0.967$\pm$0.003 &\bf 0.852$\pm$0.009\\

\multirow{-6}{*}{\textbf{Pubmed}} & \multirow{-2}{*}{Embedding} &  \textbf{ours}  & 0.861$\pm$0.001 & 0.925$\pm$0.000 &0.854$\pm$0.001& \bf 0.852$\pm$0.001 & 0.929$\pm$0.001 &\bf\bf0.849$\pm$0.005& \bf 0.858$\pm$0.002 & 0.905$\pm$0.002 &0.846$\pm$0.009\\

\midrule
               & &  Shen et al.   &  \bf  0.829$\pm$0.002 &  \bf 0.916$\pm$0.000 &\bf\bf0.831$\pm$0.003& 0.832$\pm$0.002 &  \bf 0.909$\pm$0.002 &0.829$\pm$0.005&  \bf 0.825$\pm$0.002 & \bf  0.895$\pm$0.004&\bf0.902$\pm$0.005\\

               & \multirow{-2}{*}{Prediction} &  \textbf{ours}  &  0.813$\pm$0.001 & 0.861$\pm$0.002 &0.809$\pm$0.009& \bf 0.833$\pm$0.004 & 0.873$\pm$0.000 &\bf0.833$\pm$0.009&0.824$\pm$0.004 & 0.864$\pm$0.005 &0.816$\pm$0.009\\

              \addlinespace

               & &  Shen et al.  &  0.679$\pm$0.030 & 0.711$\pm$0.000 &0.554$\pm$0.001& 0.732$\pm$0.015 & 0.769$\pm$0.014 &0.692$\pm$0.005& 0.648$\pm$0.043 & 0.680$\pm$0.039&0.658$\pm$0.022\\

               & \multirow{-2}{*}{Projection} &  \textbf{ours}  &  \bf  0.731$\pm$0.003 &  \bf 0.760$\pm$0.002 &\bf0.723$\pm$0.004& \bf 0.744$\pm$0.005 &  \bf 0.784$\pm$0.004 &\bf0.738$\pm$0.005& \bf 0.659$\pm$0.024 &  \bf 0.691$\pm$0.021 &\bf0.686$\pm$0.023\\

              \addlinespace
               & &  Shen et al.   &  \bf 0.823$\pm$0.002 &  \bf 0.893$\pm$0.000 &\bf0.823$\pm$0.001&  \bf 0.834$\pm$0.002 &  \bf 0.888$\pm$0.003 &\bf0.831$\pm$0.005&  \bf 0.822$\pm$0.001 & \bf  0.885$\pm$0.002 &\bf\bf0.892$\pm$0.001\\

\multirow{-6}{*}{\textbf{Citeseer}} & \multirow{-2}{*}{Embedding} &  \textbf{ours}  & 0.819$\pm$0.000 & 0.848$\pm$0.003 &0.819$\pm$0.009&0.831$\pm$0.003 & 0.861$\pm$0.000 &0.827$\pm$0.003&0.816$\pm$0.009 & 0.834$\pm$0.007 &0.813$\pm$0.005\\

\midrule
              & &  Shen et al.  &   \bf \bf0.926$\pm$0.001 & \bf  0.932$\pm$0.000 &\bf0.910$\pm$0.009& 0.897$\pm$0.004 & 0.910$\pm$0.001 &\bf0.882$\pm$0.003& 0.888$\pm$0.009 &  \bf 0.913$\pm$0.012&\bf0.876$\pm$0.009\\

               & \multirow{-2}{*}{Prediction} &  \textbf{ours}  &  0.920$\pm$0.001 & 0.921$\pm$0.001 &0.899$\pm$0.001& \bf 0.909$\pm$0.004 &  \bf 0.911$\pm$0.001 &0.879$\pm$0.009&\bf0.908$\pm$0.003 & 0.907$\pm$0.005 &0.863$\pm$0.003\\

              \addlinespace

               & &  Shen et al.  & 0.750$\pm$0.015 & 0.780$\pm$0.000 &0.597$\pm$0.001& 0.787$\pm$0.029 &  \bf 0.803$\pm$0.023 &\bf0.704$\pm$0.005&  \bf 0.727$\pm$0.008 & \bf  0.763$\pm$0.016&\bf0.567$\pm$0.005\\

               & \multirow{-2}{*}{Projection} &  \textbf{ours}  &   \bf 0.822$\pm$0.015 &  \bf 0.831$\pm$0.015 &\bf0.671$\pm$0.002& \bf 0.790$\pm$0.010 & 0.796$\pm$0.007 &0.648$\pm$0.009&0.663$\pm$0.026 & 0.691$\pm$0.024 &0.543$\pm$0.009\\

              \addlinespace
              & &  Shen et al.   &   \bf 0.930$\pm$0.002 &  \bf 0.925$\pm$0.000 &\bf\bf0.920$\pm$0.005&  \bf 0.915$\pm$0.001 &  \bf 0.929$\pm$0.003 &\bf0.893$\pm$0.001&  \bf 0.918$\pm$0.002 & \bf  0.931$\pm$0.005 &\bf0.904$\pm$0.002\\

\multirow{-6}{*}{\textbf{Amazon}}  & \multirow{-2}{*}{Embedding} &  \textbf{ours}  &  0.913$\pm$0.005 & 0.912$\pm$0.004 &0.887$\pm$0.003&0.875$\pm$0.014 & 0.894$\pm$0.011 &0.844$\pm$0.001&0.907$\pm$0.005 & 0.904$\pm$0.005 &0.876$\pm$0.001\\

\midrule
               & &  Shen et al.  &   \bf 0.945$\pm$0.001 & \bf  0.971$\pm$0.000 &\bf0.924$\pm$0.002& 0.904$\pm$0.006 & 0.929$\pm$0.006 &0.891$\pm$0.009&  \bf 0.943$\pm$0.001 & \bf  0.974$\pm$0.002&0.925$\pm$0.004\\

              & \multirow{-2}{*}{Prediction} &  \textbf{ours}  &  0.942$\pm$0.004 & 0.962$\pm$0.005 &0.922$\pm$0.001& \bf 0.934$\pm$0.003 & \bf  0.956$\pm$0.005 &\bf0.901$\pm$0.004&0.941$\pm$0.000 & 0.968$\pm$0.001 &\bf0.926$\pm$0.009\\

              \addlinespace

               & &  Shen et al.   &  0.833$\pm$0.029 & 0.849$\pm$0.000 &0.553$\pm$0.004& 0.822$\pm$0.028 & 0.845$\pm$0.030 &0.684$\pm$0.001& 0.755$\pm$0.021 & 0.776$\pm$0.017&0.456$\pm$0.031\\

              & \multirow{-2}{*}{Projection} &  \textbf{ours}  &  \bf  0.927$\pm$0.000 & \bf  0.949$\pm$0.001 &\bf0.900$\pm$0.001& \bf 0.908$\pm$0.006 &  \bf 0.930$\pm$0.006 &\bf0.860$\pm$0.005& \bf 0.905$\pm$0.006 & \bf  0.927$\pm$0.007 &\bf0.863$\pm$0.002\\

              \addlinespace
              & &  Shen et al.   & 0.940$\pm$0.001 & 0.964$\pm$0.000 &0.918$\pm$0.001& 0.886$\pm$0.008 & 0.902$\pm$0.009 &0.901$\pm$0.009& 0.939$\pm$0.001 & 0.969$\pm$0.002 &0.921$\pm$0.002\\

\multirow{-6}{*}{\textbf{Coauthor}}  & \multirow{-2}{*}{Embedding} &  \textbf{ours}  &   \bf 0.942$\pm$0.002 & \bf  0.959$\pm$0.003 &\bf0.926$\pm$0.001& \bf 0.942$\pm$0.002 &  \bf 0.954$\pm$0.004 &\bf0.918$\pm$0.005& \bf 0.940$\pm$0.003 &  \bf 0.950$\pm$0.003 &\bf0.923$\pm$0.004\\

\midrule
             & &  Shen et al.   &   \bf 0.898$\pm$0.003 &  \bf 0.933$\pm$0.000 &\bf\bf0.899$\pm$0.002& 0.891$\pm$0.003 & 0.941$\pm$0.003 &0.889$\pm$0.003& \bf  0.895$\pm$0.003 & \bf  0.935$\pm$0.003&0.883$\pm$0.005\\

               & \multirow{-2}{*}{Prediction} &  \textbf{ours}  &  0.888$\pm$0.003 & 0.903$\pm$0.006 &0.887$\pm$0.005&\bf0.892$\pm$0.003 &  \bf 0.923$\pm$0.017 &\bf0.896$\pm$0.009&0.888$\pm$0.004 & 0.914$\pm$0.004 &\bf\bf0.892$\pm$0.001\\

              \addlinespace

              & &  Shen et al.   & 0.849$\pm$0.047 & 0.884$\pm$0.100 &0.887$\pm$0.003& 0.869$\pm$0.002 & 0.913$\pm$0.002 &0.863$\pm$0.002& \bf  0.891$\pm$0.005 &  \bf 0.931$\pm$0.008&\bf0.888$\pm$0.001\\

             & \multirow{-2}{*}{Projection} &  \textbf{ours}  &  \bf  0.887$\pm$0.004 & \bf  0.912$\pm$0.003 &\bf0.892$\pm$0.004& \bf 0.887$\pm$0.008 & \bf  0.919$\pm$0.010 &\bf0.892$\pm$0.004&0.865$\pm$0.014 & 0.887$\pm$0.017 &0.887$\pm$0.001\\

              \addlinespace
               & &  Shen et al.   &   \bf 0.902$\pm$0.002 & \bf  0.936$\pm$0.000 &\bf0.886$\pm$0.004& 0.876$\pm$0.001 & \bf  0.925$\pm$0.002 &0.866$\pm$0.001& 0.861$\pm$0.014 &  \bf 0.889$\pm$0.009 &0.852$\pm$0.002\\

\multirow{-6}{*}{\textbf{ACM}} & \multirow{-2}{*}{Embedding} &  \textbf{ours}  &  0.883$\pm$0.007 & 0.891$\pm$0.010 &0.873$\pm$0.001& \bf 0.882$\pm$0.008 & 0.911$\pm$0.015 &\bf0.886$\pm$0.001& \bf 0.870$\pm$0.007 & 0.874$\pm$0.013 &\bf0.870$\pm$0.004\\
\bottomrule
\end{tabular}
}
\label{tab:acc_fidelity_type1_sage}
\end{table*}

\clearpage

\section{Additional result - Type II attack}\label{sec:type2_performance_appendix}
\begin{table*}[!h]
\centering
\caption{
The accuracy and fidelity scores of \textbf{Type II} attacks on all of the datasets with \textbf{GAT} as the target model. }
\vspace{0.5cm}
\scalebox{0.8}{
\begin{tabular}{lclccccccccc}
\toprule
\multirow{3}{*}{\textbf{Dataset}} &
  \multirow{3}{*}{\begin{tabular}[x]{@{}c@{}} \textbf{Task} \end{tabular}} 
& & 
  \multicolumn{3}{c}{\textbf{Surrogate GIN}} & 
  \multicolumn{3}{c}{\textbf{Surrogate GAT}} & 
  \multicolumn{3}{c}{\textbf{Surrogate SAGE}} \\
 \cmidrule(lr){4-6} \cmidrule(lr){7-9} \cmidrule(lr){10-12}
 & &
   &
  \textbf{Accuracy} &
  \textbf{Fidelity} &
  \textbf{F1 score} &
  \textbf{Accuracy} &
  \textbf{Fidelity} &
  \textbf{F1 score} &
  \textbf{Accuracy} &
  \textbf{Fidelity} &
  \textbf{F1 score}\\

\midrule
             & &  Shen et al.   &0.736$\pm$0.000 & \bf0.855$\pm$0.000 &\bf\bf0.682$\pm$0.002& 0.733$\pm$0.001 &\bf 0.845$\pm$0.000 &0.617$\pm$0.002& 0.747$\pm$0.000 & \bf0.856$\pm$0.000&0.610$\pm$0.002\\

               & \multirow{-2}{*}{Prediction} &  \textbf{ours}  & \bf 0.745$\pm$0.004 & 0.811$\pm$0.001 &0.631$\pm$0.004&\bf0.742$\pm$0.001 & 0.820$\pm$0.003 &\bf0.637$\pm$0.002&\bf0.749$\pm$0.002 & 0.815$\pm$0.002 &\bf0.620$\pm$0.009\\

              \addlinespace

              & &  Shen et al.  & 0.668$\pm$0.000 & 0.746$\pm$0.000 &0.483$\pm$0.002& 0.704$\pm$0.002 &\bf 0.794$\pm$0.002 &0.545$\pm$0.002& 0.636$\pm$0.002 & 0.697$\pm$0.003&0.403$\pm$0.001\\

              & \multirow{-2}{*}{Projection} &  \textbf{ours}  & \bf 0.701$\pm$0.003 &\bf 0.756$\pm$0.004 &\bf\bf0.550$\pm$0.005&\bf0.715$\pm$0.003 & 0.775$\pm$0.004 &\bf0.571$\pm$0.005&\bf0.688$\pm$0.004 &\bf 0.742$\pm$0.006 &\bf0.501$\pm$0.003\\

              \addlinespace
               & &  Shen et al.   &0.733$\pm$0.001 & \bf0.837$\pm$0.000 &0.637$\pm$0.002& \bf0.746$\pm$0.000 & \bf0.858$\pm$0.000 &\bf0.677$\pm$0.001& \bf0.748$\pm$0.005 &\bf 0.861$\pm$0.010  &0.626$\pm$0.003\\

\multirow{-6}{*}{\textbf{DBLP}} & \multirow{-2}{*}{Embedding} &  \textbf{ours}  &  \bf0.736$\pm$0.004 & 0.790$\pm$0.003 &\bf\bf0.663$\pm$0.002&0.745$\pm$0.008 & 0.778$\pm$0.015 &0.672$\pm$0.004&0.705$\pm$0.001 & 0.742$\pm$0.002 &\bf0.634$\pm$0.001\\

\midrule
               & &  Shen et al.   &0.835$\pm$0.000 & \bf0.911$\pm$0.000 &0.826$\pm$0.001& 0.817$\pm$0.001 & \bf0.910$\pm$0.007 &0.818$\pm$0.003& 0.832$\pm$0.000 & \bf0.913$\pm$0.000&0.827$\pm$0.001\\

              & \multirow{-2}{*}{Prediction} &  \textbf{ours}  &  0\bf.850$\pm$0.000 & 0.875$\pm$0.001 &\bf\bf0.844$\pm$0.001&\bf0.826$\pm$0.007 & 0.880$\pm$0.007 &\bf0.821$\pm$0.005&\bf0.849$\pm$0.001 & 0.875$\pm$0.001 &\bf0.844$\pm$0.009\\

              \addlinespace

              & &  Shen et al.   & 0.749$\pm$0.000 & 0.800$\pm$0.000 &0.720$\pm$0.004& 0.797$\pm$0.001 &\bf 0.892$\pm$0.001 &\bf0.813$\pm$0.002& 0.751$\pm$0.000 & 0.806$\pm$0.000&0.728$\pm$0.003\\

              & \multirow{-2}{*}{Projection} &  \textbf{ours}  &\bf  \bf0.838$\pm$0.002 &\bf 0.867$\pm$0.001 &\bf\bf0.830$\pm$0.004&\bf0.826$\pm$0.005 & 0.881$\pm$0.005 &0.805$\pm$0.009&\bf0.837$\pm$0.000 &\bf 0.860$\pm$0.001 &\bf0.828$\pm$0.004\\

              \addlinespace
               & &  Shen et al.  & 0.844$\pm$0.000 & \bf0.911$\pm$0.000 &0.837$\pm$0.009& 0.827$\pm$0.000 & \bf0.911$\pm$0.000 &0.822$\pm$0.005& 0.838$\pm$0.001 &\bf 0.931$\pm$0.00 &0.833$\pm$0.002\\

\multirow{-6}{*}{\textbf{Pubmed}} & \multirow{-2}{*}{Embedding} &  \textbf{ours}  & \bf 0.851$\pm$0.002 & 0.867$\pm$0.002 &\bf\bf0.851$\pm$0.001&\bf0.841$\pm$0.000 & 0.874$\pm$0.001 &\bf0.831$\pm$0.003&\bf0.845$\pm$0.002 & 0.847$\pm$0.002 &\bf0.851$\pm$0.004\\

\midrule
              & &  Shen et al.   &\bf 0.822$\pm$0.001 & \bf0.902$\pm$0.000 &0.816$\pm$0.005& 0.835$\pm$0.001 &\bf 0.893$\pm$0.001 &0.835$\pm$0.001& \bf0.828$\pm$0.000 & \bf0.895$\pm$0.001&\bf0.829$\pm$0.004\\

               & \multirow{-2}{*}{Prediction} &  \textbf{ours}  &  0.817$\pm$0.004 & 0.849$\pm$0.002 &\bf\bf0.818$\pm$0.001&\bf0.843$\pm$0.003 & 0.880$\pm$0.002 &0.835$\pm$0.009&0.824$\pm$0.004 & 0.862$\pm$0.001 &0.817$\pm$0.001\\

              \addlinespace

              & &  Shen et al.  &  0.682$\pm$0.008 & 0.713$\pm$0.000 &0.671$\pm$0.009& 0.742$\pm$0.003 &\bf 0.787$\pm$0.002 &0.721$\pm$0.001& 0.624$\pm$0.021 & 0.661$\pm$0.023&0.655$\pm$0.005\\

               & \multirow{-2}{*}{Projection} &  \textbf{ours}  & \bf \bf0.730$\pm$0.001 & \bf0.749$\pm$0.001 &\bf0.821$\pm$0.002&\bf0.757$\pm$0.001 & 0.783$\pm$0.001 &\bf0.736$\pm$0.005&\bf0.695$\pm$0.007 &\bf 0.712$\pm$0.001 &\bf0.663$\pm$0.001\\

              \addlinespace
               & &  Shen et al.   & \bf 0.825$\pm$0.001 & \bf0.909$\pm$0.000 &\bf0.824$\pm$0.009& 0.833$\pm$0.001 & \bf0.918$\pm$0.002 &0.847$\pm$0.002& \bf0.819$\pm$0.001 &\bf 0.892$\pm$0.001 &0.821$\pm$0.005\\

\multirow{-6}{*}{\textbf{Citeseer}} & \multirow{-2}{*}{Embedding} &  \textbf{ours}  &  0.814$\pm$0.006 & 0.843$\pm$0.005 &0.820$\pm$0.002&\bf\bf0.844$\pm$0.001 & 0.869$\pm$0.006 &\bf0.857$\pm$0.009&0.812$\pm$0.005 & 0.820$\pm$0.005 &\bf0.829$\pm$0.002\\

\midrule
              & &  Shen et al.  & 0.907$\pm$0.000 & 0.922$\pm$0.000 &0.877$\pm$0.003&\bf 0.905$\pm$0.002 &\bf 0.930$\pm$0.004 &0.775$\pm$0.009& \bf0.886$\pm$0.003 &\bf 0.906$\pm$0.004&0.829$\pm$0.001\\

              & \multirow{-2}{*}{Prediction} &  \textbf{ours}  &\bf  0.922$\pm$0.001 & \bf0.932$\pm$0.004 &\bf0.905$\pm$0.001&0.896$\pm$0.003 & 0.925$\pm$0.004 &\bf0.902$\pm$0.005&0.879$\pm$0.011 & 0.895$\pm$0.011 &\bf0.868$\pm$0.001\\

              \addlinespace

               & &  Shen et al.  & 0.446$\pm$0.002 & 0.444$\pm$0.000 &0.368$\pm$0.009& 0.395$\pm$0.091 & 0.396$\pm$0.086 &0.273$\pm$0.002& \bf0.500$\pm$0.002 & \bf0.495$\pm$0.002&0.298$\pm$0.009\\

               & \multirow{-2}{*}{Projection} &  \textbf{ours}  & \bf 0.500$\pm$0.006 & \bf0.490$\pm$0.006 &\bf0.389$\pm$0.009&\bf0.497$\pm$0.008 & \bf0.492$\pm$0.008 &\bf0.387$\pm$0.001&0.400$\pm$0.075 & 0.399$\pm$0.071 &\bf0.301$\pm$0.003\\

              \addlinespace
               & &  Shen et al.   & 0.750$\pm$0.011 & 0.766$\pm$0.000 &\bf0.889$\pm$0.005&\bf 0.864$\pm$0.026 & 0.878$\pm$0.032 &0.872$\pm$0.004& 0.589$\pm$0.026 & 0.606$\pm$0.025  &0.424$\pm$0.001\\

\multirow{-6}{*}{\textbf{Amazon}}  & \multirow{-2}{*}{Embedding} &  \textbf{ours}  & \bf 0.906$\pm$0.005 &\bf 0.924$\pm$0.003 &0.874$\pm$0.004&0.861$\pm$0.032 & \bf0.882$\pm$0.034 &\bf0.887$\pm$0.004&\bf0.839$\pm$0.011 & \bf0.851$\pm$0.012 &\bf0.765$\pm$0.009\\

\midrule
               & &  Shen et al.  &  0.948$\pm$ 0.001 & \bf 0.960$\pm$ 0.002 &0.943$\pm$0.001& \bf 0.906$\pm$ 0.001 & \bf0.927$\pm$ 0.002 &0.911$\pm$0.003&  0.939$\pm$ 0.003 &  0.953$\pm$ 0.002 &0.909$\pm$0.001\\

              & \multirow{-2}{*}{Prediction} &  \textbf{ours}  & 0.948$\pm$0.002 & 0.959$\pm$0.003 &\bf0.944$\pm$0.001&0.895$\pm$0.006 & 0.911$\pm$0.007 &\bf0.918$\pm$0.005&\bf\bf0.944$\pm$0.001 &\bf 0.956$\pm$0.001 &\bf0.921$\pm$0.009\\
              
              \addlinespace

               & &  Shen et al.   &\bf  0.821$\pm$ 0.001 &  \bf0.827$\pm$ 0.001 &0.810$\pm$0.003& \bf 0.859$\pm$ 0.002 &\bf 0.937$\pm$ 0.002 &\bf0.850$\pm$0.001& \bf 0.799$\pm$ 0.021 & \bf 0.850$\pm$ 0.031 &\bf0.720$\pm$0.003\\

              & \multirow{-2}{*}{Projection} &  \textbf{ours}  &  0.789$\pm$0.006 & 0.795$\pm$0.007 &\bf0.811$\pm$0.005&0.673$\pm$0.020 & 0.684$\pm$0.021 &0.765$\pm$0.002&0.758$\pm$0.009 & 0.765$\pm$0.008 &0.710$\pm$0.001\\

              \addlinespace
              & &  Shen et al.   &\bf  0.945$\pm$ 0.001 &  0.826$\pm$ 0.001 &\bf\bf\bf0.944$\pm$0.004& \bf 0.920$\pm$ 0.002 & \bf 0.937$\pm$ 0.001 &0.910$\pm$0.004&  0.930$\pm$ 0.007 &  0.947$\pm$ 0.011 &0.910$\pm$0.002\\

\multirow{-6}{*}{\textbf{Coauthor}}  & \multirow{-2}{*}{Embedding} &  \textbf{ours}  &  0.941$\pm$0.002 & \bf0.952$\pm$0.001 &0.940$\pm$0.002&0.904$\pm$0.009 & 0.921$\pm$0.008 &\bf\bf0.920$\pm$0.009&\bf0.939$\pm$0.001 &\bf 0.948$\pm$0.001 &\bf\bf0.920$\pm$0.009\\

\midrule
              & &  Shen et al.   &  0.898$\pm$0.000 & \bf0.947$\pm$0.000 &0.891$\pm$0.003&\bf 0.900$\pm$0.000 &\bf 0.958$\pm$0.000 &\bf0.886$\pm$0.009& \bf0.892$\pm$0.000 &\bf 0.936$\pm$0.000&0.889$\pm$0.005\\

               & \multirow{-2}{*}{Prediction} &  \textbf{ours}  & \bf 0.903$\pm$0.004 & 0.923$\pm$0.005 &\bf0.903$\pm$0.009&0.896$\pm$0.007 & 0.940$\pm$0.003 &0.876$\pm$0.001&0.889$\pm$0.012 & 0.923$\pm$0.007 &\bf\bf\bf0.902$\pm$0.002\\

              \addlinespace

              & &  Shen et al.   &  0.876$\pm$0.000 & \bf0.915$\pm$0.000 &0.869$\pm$0.001&\bf 0.857$\pm$0.000 &\bf 0.913$\pm$0.000 &\bf0.859$\pm$0.009& \bf0.888$\pm$0.000 & \bf0.929$\pm$0.000&\bf0.886$\pm$0.009\\

              & \multirow{-2}{*}{Projection} &  \textbf{ours}  &\bf 0.880$\pm$0.002 & 0.906$\pm$0.002 &\bf0.889$\pm$0.002&0.842$\pm$0.012 & 0.891$\pm$0.010 &0.842$\pm$0.003&0.867$\pm$0.003 & 0.886$\pm$0.009 &0.868$\pm$0.004\\

              \addlinespace
               & &  Shen et al.   & \bf 0.900$\pm$0.000 & \bf0.927$\pm$0.000 &\bf0.897$\pm$0.005& 0.845$\pm$0.000 & 0.880$\pm$0.000 &\bf 0.892$\pm$0.001& 0.839$\pm$0.000 & 0.855$\pm$0.000  &0.840$\pm$0.009\\

\multirow{-6}{*}{\textbf{ACM}} & \multirow{-2}{*}{Embedding} &  \textbf{ours}  &  0.835$\pm$0.010 & 0.848$\pm$0.016 & 0.883$\pm$0.001&\bf0.864$\pm$0.013 & \bf0.900$\pm$0.011 &0.877$\pm$0.003&\bf0.853$\pm$0.003 &\bf 0.862$\pm$0.007 &\bf0.855$\pm$0.003\\
\bottomrule
\end{tabular}
}
\label{tab:acc_fidelity_typei1_gat}
\end{table*}

\clearpage

\begin{table*}[!t]
\centering
\caption{
The accuracy and fidelity scores of Type II attacks on all of the datasets with SAGE as the target model. }
\vspace{0.5cm}

\scalebox{0.8}{
\begin{tabular}{lclccccccccc}
\toprule
\multirow{3}{*}{\textbf{Dataset}} &
  \multirow{3}{*}{\begin{tabular}[x]{@{}c@{}} \textbf{Task} \end{tabular}} 
& & 
  \multicolumn{3}{c}{\textbf{Surrogate GIN}} & 
  \multicolumn{3}{c}{\textbf{Surrogate GAT}} & 
  \multicolumn{3}{c}{\textbf{Surrogate SAGE}} \\
 \cmidrule(lr){4-6} \cmidrule(lr){7-9} \cmidrule(lr){10-12}
 & &
   &
  \textbf{Accuracy} &
  \textbf{Fidelity} &
  \textbf{F1 score} &
  \textbf{Accuracy} &
  \textbf{Fidelity} &
  \textbf{F1 score} &
  \textbf{Accuracy} &
  \textbf{Fidelity} &
  \textbf{F1 score}\\

\midrule
             & &  Shen et al.   & 0.761$\pm$0.000 & \bf  0.904$\pm$0.000 &0.682$\pm$0.001& 0.759$\pm$0.000 &  \bf 0.864$\pm$0.000 &0.689$\pm$0.001&  \bf 0.764$\pm$0.000 & \bf  0.891$\pm$0.001&\bf0.683$\pm$0.001\\
             
               & \multirow{-2}{*}{Prediction} &  \textbf{ours}  &  0.761$\pm$0.002 & 0.858$\pm$0.002 &\bf0.696$\pm$0.001& \bf 0.770$\pm$0.007 & 0.853$\pm$0.005 &\bf0.690$\pm$0.001&0.760$\pm$0.005 & 0.849$\pm$0.008 &0.656$\pm$0.001\\

              \addlinespace

              & &  Shen et al.  & 0.714$\pm$0.001 & 0.807$\pm$0.000 &0.483$\pm$0.001& 0.734$\pm$0.000 &  \bf 0.839$\pm$0.000 &0.596$\pm$0.001& 0.729$\pm$0.001 &  \bf 0.830$\pm$0.000&0.403$\pm$0.001\\

              & \multirow{-2}{*}{Projection} &  \textbf{ours}  &   \bf 0.731$\pm$0.002 &  \bf 0.821$\pm$0.003 &\bf0.570$\pm$0.004& \bf 0.737$\pm$0.004 & 0.824$\pm$0.006 &\bf0.599$\pm$0.005& \bf 0.743$\pm$0.000 & 0.829$\pm$0.001 &\bf0.545$\pm$0.001\\

              \addlinespace
               & &  Shen et al.   & \bf   0.757$\pm$0.000 &  \bf 0.887$\pm$0.000 &\bf0.637$\pm$0.004& \bf  0.776$\pm$0.001 & \bf  0.885$\pm$0.001 &\bf0.710$\pm$0.009&  \bf 0.762$\pm$0.000 &  \bf 0.895$\pm$0.000 &0.626$\pm$0.002\\

\multirow{-6}{*}{\textbf{DBLP}} & \multirow{-2}{*}{Embedding} &  \textbf{ours}  &  0.737$\pm$0.017 & 0.809$\pm$0.021 &0.632$\pm$0.002&0.764$\pm$0.003 & 0.837$\pm$0.003 &0.701$\pm$0.009&0.724$\pm$0.003 & 0.787$\pm$0.003 &\bf0.645$\pm$0.002\\

\midrule
               & &  Shen et al.   & 0.855$\pm$0.000 & \bf 0.956$\pm$0.000 &\bf0.851$\pm$0.004&   0.839$\pm$0.000 &  \bf 0.944$\pm$0.001 &\bf0.840$\pm$0.001&  \bf 0.856$\pm$0.000 & \bf  0.952$\pm$0.001&0.847$\pm$0.001\\

              & \multirow{-2}{*}{Prediction} &  \textbf{ours}  &  \bf  0.856$\pm$0.002 & 0.925$\pm$0.001 &0.850$\pm$0.003& \bf 0.846$\pm$0.003 & 0.925$\pm$0.003 &0.826$\pm$0.009&0.855$\pm$0.001 & 0.924$\pm$0.000 &\bf0.849$\pm$0.004\\

              \addlinespace

              & &  Shen et al.   &  \bf 0.852$\pm$0.000 &  \bf 0.951$\pm$0.000 &0.820$\pm$0.003& 0.822$\pm$0.001 &  \bf 0.922$\pm$0.001 &0.826$\pm$0.005&  \bf 0.845$\pm$0.000 &  \bf 0.938$\pm$0.000&0.823$\pm$0.001\\

              & \multirow{-2}{*}{Projection} &  \textbf{ours}  &  0.848$\pm$0.001 & 0.920$\pm$0.001 &\bf\bf0.844$\pm$0.001& \bf \bf0.838$\pm$0.001 & 0.920$\pm$0.002 &\bf0.830$\pm$0.001&0.842$\pm$0.001 & 0.915$\pm$0.002 &\bf0.834$\pm$0.001\\

              \addlinespace
               & &  Shen et al.  &  \bf  0.859$\pm$0.001 & 0.958$\pm$0.000 &0.853$\pm$0.003&   0.842$\pm$0.002 & \bf  0.944$\pm$0.003 &0.838$\pm$0.001&  \bf 0.848$\pm$0.000 &  \bf 0.953$\pm$0.000  &0.847$\pm$0.001\\

\multirow{-6}{*}{\textbf{Pubmed}} & \multirow{-2}{*}{Embedding} &  \textbf{ours}  &  0.859$\pm$0.000 &  \bf 0.920$\pm$0.005 &\bf0.854$\pm$0.001&\bf0.848$\pm$0.002 & 0.924$\pm$0.002 &\bf\bf0.844$\pm$0.001&0.847$\pm$0.003 & 0.895$\pm$0.006 &0.847$\pm$0.001\\

\midrule
               & &  Shen et al.   &   \bf 0.829$\pm$0.001 &  \bf 0.933$\pm$0.000 &\bf0.832$\pm$0.005&  \bf 0.846$\pm$0.001 &  \bf 0.908$\pm$0.000 &\bf0.838$\pm$0.001& \bf  0.831$\pm$0.000 &  \bf 0.908$\pm$0.000&\bf0.832$\pm$0.009\\

               & \multirow{-2}{*}{Prediction} &  \textbf{ours}  &  0.814$\pm$0.003 & 0.867$\pm$0.001 &0.819$\pm$0.001&0.843$\pm$0.004 & 0.897$\pm$0.003 &0.837$\pm$0.004&0.818$\pm$0.001 & 0.876$\pm$0.006 &0.809$\pm$0.004\\

              \addlinespace

               & &  Shen et al.  &  0.685$\pm$0.004 & 0.727$\pm$0.000 &0.617$\pm$0.001& 0.736$\pm$0.001 & 0.792$\pm$0.003 &0.734$\pm$0.002&  \bf 0.719$\pm$0.003 &  \bf 0.756$\pm$0.004&0.602$\pm$0.005\\

               & \multirow{-2}{*}{Projection} &  \textbf{ours}  &  \bf  0.718$\pm$0.004 & \bf  0.762$\pm$0.005 &\bf0.707$\pm$0.003& \bf 0.750$\pm$0.003 &  \bf 0.797$\pm$0.001 &0.734$\pm$0.001&0.666$\pm$0.015 & 0.704$\pm$0.014 &\bf0.655$\pm$0.005\\

              \addlinespace
               & &  Shen et al.   &  \bf  \bf0.830$\pm$0.001 &  \bf 0.913$\pm$0.000 &\bf0.825$\pm$0.004& 0.840$\pm$0.001 &  \bf 0.905$\pm$0.001 &0.829$\pm$0.003&  \bf 0.820$\pm$0.002 &  \bf 0.886$\pm$0.001 &0.817$\pm$0.001\\

\multirow{-6}{*}{\textbf{Citeseer}} & \multirow{-2}{*}{Embedding} &  \textbf{ours}  &  0.812$\pm$0.005 & 0.849$\pm$0.004 &0.815$\pm$0.001& \bf 0.845$\pm$0.002 & 0.878$\pm$0.001 &\bf0.841$\pm$0.001&0.803$\pm$0.004 & 0.824$\pm$0.002 &\bf0.812$\pm$0.004\\

\midrule
              & &  Shen et al.  &  \bf  0.924$\pm$0.001 &  \bf 0.953$\pm$0.000 &0.914$\pm$0.005& 0.899$\pm$0.009 &  \bf 0.932$\pm$0.007 &0.702$\pm$0.001& 0.876$\pm$0.002 &  \bf 0.921$\pm$0.002&0.830$\pm$0.003\\

               & \multirow{-2}{*}{Prediction} &  \textbf{ours}  &  0.922$\pm$0.005 & 0.932$\pm$0.009 &\bf0.916$\pm$0.009& \bf 0.909$\pm$0.004 & 0.930$\pm$0.008 &\bf0.857$\pm$0.005& \bf 0.899$\pm$0.008 & 0.913$\pm$0.011 &\bf0.867$\pm$0.003\\

              \addlinespace

               & &  Shen et al.  &  0.720$\pm$0.004 & 0.743$\pm$0.000 &0.571$\pm$0.004& 0.767$\pm$0.011 &  \bf 0.818$\pm$0.008 &0.338$\pm$0.002& 0.638$\pm$0.002 &  \bf 0.674$\pm$0.002&0.554$\pm$0.001\\

               & \multirow{-2}{*}{Projection} &  \textbf{ours}  &  \bf  0.856$\pm$0.013 & \bf  0.854$\pm$0.014 &\bf0.820$\pm$0.009& \bf 0.779$\pm$0.010 & 0.794$\pm$0.004 &\bf0.515$\pm$0.002& \bf 0.654$\pm$0.028 & 0.670$\pm$0.033 &0.554$\pm$0.001\\

              \addlinespace
              & &  Shen et al.   &  \bf  0.926$\pm$0.001 &  \bf 0.948$\pm$0.000 &\bf0.914$\pm$0.002&  \bf 0.900$\pm$0.007 &  \bf 0.939$\pm$0.005 &0.703$\pm$0.004& 0.859$\pm$0.001 & 0.870$\pm$0.001 &\bf0.901$\pm$0.009\\

\multirow{-6}{*}{\textbf{Amazon}}  & \multirow{-2}{*}{Embedding} &  \textbf{ours}  &  0.921$\pm$0.003 & 0.929$\pm$0.003 &0.909$\pm$0.005&0.894$\pm$0.018 & 0.916$\pm$0.035 &\bf0.889$\pm$0.005& \bf 0.906$\pm$0.004 & \bf  0.913$\pm$0.005 &0.860$\pm$0.001\\

\midrule
               & &  Shen et al.  &  \bf 0.948$\pm$0.000 &  \bf 0.979$\pm$0.000 &0.960$\pm$0.009& 0.866$\pm$0.004 & 0.891$\pm$0.004 &0.874$\pm$0.001& 0.945$\pm$0.000 & \bf  0.981$\pm$0.000&0.971$\pm$0.005\\

              & \multirow{-2}{*}{Prediction} &  \textbf{ours}  &  0.946$\pm$0.001 & 0.967$\pm$0.001 &\bf0.961$\pm$0.003& \bf 0.884$\pm$0.040 & \bf  0.909$\pm$0.042 &\bf0.898$\pm$0.002& \bf 0.947$\pm$0.002 & 0.971$\pm$0.002 &\bf0.973$\pm$0.005\\
              
              \addlinespace

               & &  Shen et al.   & 0.731$\pm$0.001 & 0.743$\pm$0.000 &0.714$\pm$0.005& \bf  0.844$\pm$0.007 & \bf  0.865$\pm$0.008 &\bf0.820$\pm$0.001& 0.717$\pm$0.005 & 0.729$\pm$0.005&0.704$\pm$0.004\\
              & \multirow{-2}{*}{Projection} &  \textbf{ours}  &   \bf 0.907$\pm$0.011 &  \bf 0.924$\pm$0.011 &\bf0.899$\pm$0.004&0.674$\pm$0.053 & 0.682$\pm$0.054 &0.750$\pm$0.001& \bf 0.782$\pm$0.003 &  \bf 0.802$\pm$0.005 &\bf0.790$\pm$0.001\\

              \addlinespace
              & &  Shen et al.   &  \bf  0.947$\pm$0.000 &  \bf 0.977$\pm$0.000 &0.959$\pm$0.004& 0.878$\pm$0.002 & 0.894$\pm$0.006 &0.873$\pm$0.001& 0.890$\pm$0.030 & 0.905$\pm$0.034  &0.899$\pm$0.001\\

\multirow{-6}{*}{\textbf{Coauthor}}  & \multirow{-2}{*}{Embedding} &  \textbf{ours}  &  0.938$\pm$0.003 & 0.952$\pm$0.006 &\bf\bf0.960$\pm$0.009& \bf 0.921$\pm$0.004 &  \bf 0.940$\pm$0.004 &\bf0.930$\pm$0.003& \bf 0.938$\pm$0.001 & \bf  0.954$\pm$0.003 &\bf0.930$\pm$0.001\\

\midrule
              & &  Shen et al.   &  \bf 0.907$\pm$0.000 &  \bf 0.950$\pm$0.000 &0.896$\pm$0.001&  \bf 0.902$\pm$0.000 &  \bf 0.950$\pm$0.000 &\bf0.888$\pm$0.002&  \bf 0.905$\pm$0.000 &  \bf 0.943$\pm$0.000&\bf0.893$\pm$0.001\\

                & \multirow{-2}{*}{Prediction} &  \textbf{ours}  &  0.905$\pm$0.003 & 0.924$\pm$0.002 &\bf0.910$\pm$0.009&0.900$\pm$0.006 & 0.938$\pm$0.002 &0.884$\pm$0.001&0.889$\pm$0.006 & 0.910$\pm$0.012 &0.890$\pm$0.001\\

              \addlinespace

              & &  Shen et al.   &   \bf 0.901$\pm$0.000 & \bf  0.938$\pm$0.000 &0.888$\pm$0.004& \bf  0.896$\pm$0.000 &  \bf 0.944$\pm$0.000 &\bf0.891$\pm$0.002&  \bf 0.904$\pm$0.000 & \bf  0.939$\pm$0.000&0.881$\pm$0.002\\

              & \multirow{-2}{*}{Projection} &  \textbf{ours}  &  0.895$\pm$0.002 & 0.915$\pm$0.002 &\bf0.902$\pm$0.004&0.887$\pm$0.003 & 0.920$\pm$0.004 &0.889$\pm$0.001&0.869$\pm$0.004 & 0.886$\pm$0.003 &\bf0.836$\pm$0.009\\

              \addlinespace
               & &  Shen et al.   &  \bf 0.907$\pm$0.000 &  \bf 0.952$\pm$0.000 &\bf0.905$\pm$0.009&  \bf 0.868$\pm$0.000 & \bf  0.919$\pm$0.000 &\bf0.881$\pm$0.001& 0.870$\pm$0.000 & 0.895$\pm$0.000 &0.867$\pm$0.005\\

\multirow{-6}{*}{\textbf{ACM}} & \multirow{-2}{*}{Embedding} &  \textbf{ours}  &  0.879$\pm$0.004 & 0.890$\pm$0.007 &0.898$\pm$0.009&0.841$\pm$0.027 & 0.836$\pm$0.033 &0.878$\pm$0.009& \bf 0.877$\pm$0.007 & \bf  0.893$\pm$0.006 &\bf0.871$\pm$0.009\\
\bottomrule
\end{tabular}
}
\label{tab:acc_fidelity_type1i_sage}
\end{table*}

\clearpage

\section{Additional results - query budget}\label{sec:appendix_query}
\begin{figure}[h!]
\centering
\includegraphics[trim={2.5cm 2cm 2cm 2.3cm},clip,width=0.7\linewidth]{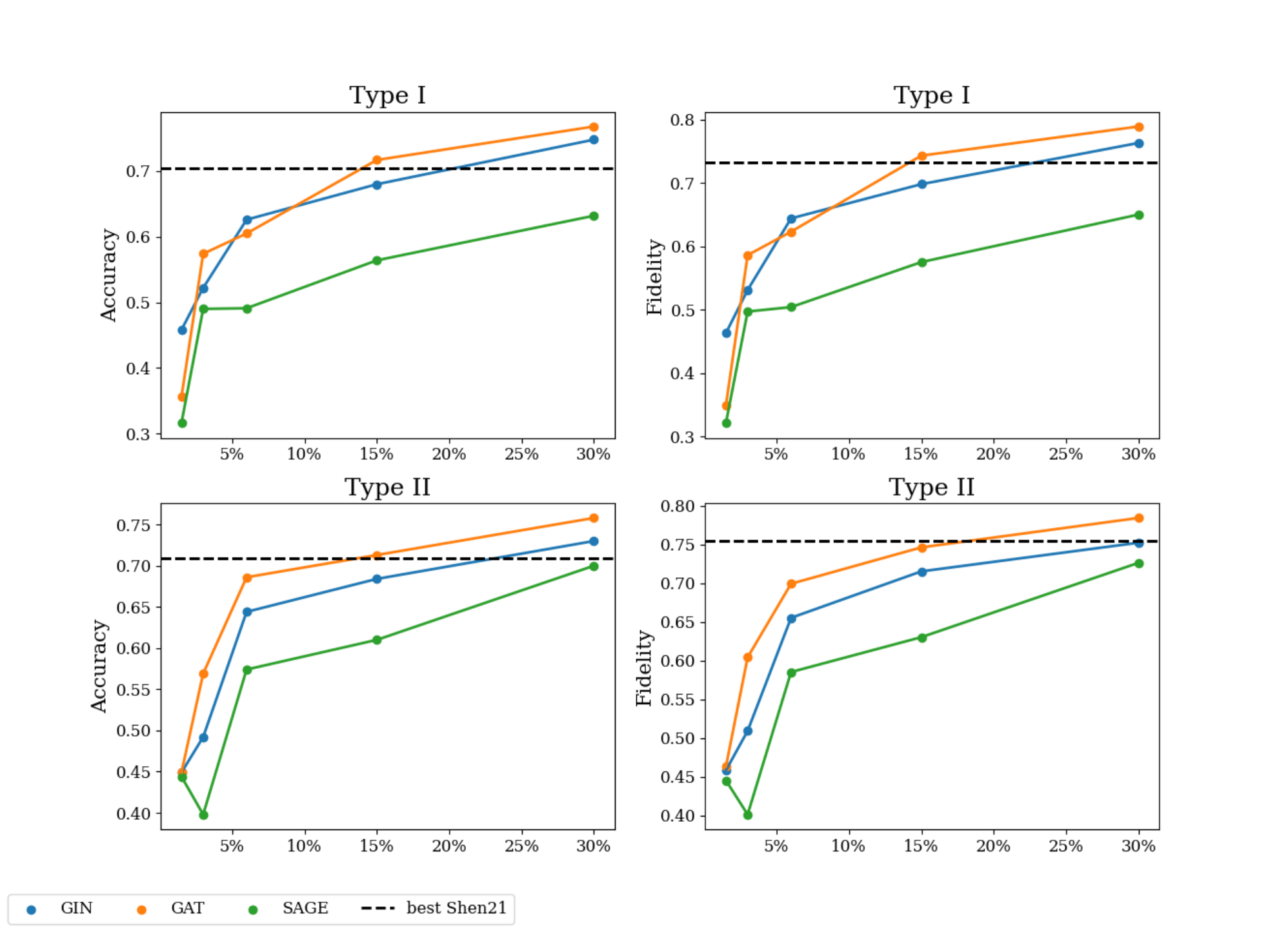} 
\includegraphics[trim={0 0cm 20cm 23cm},clip,width=0.5\linewidth]{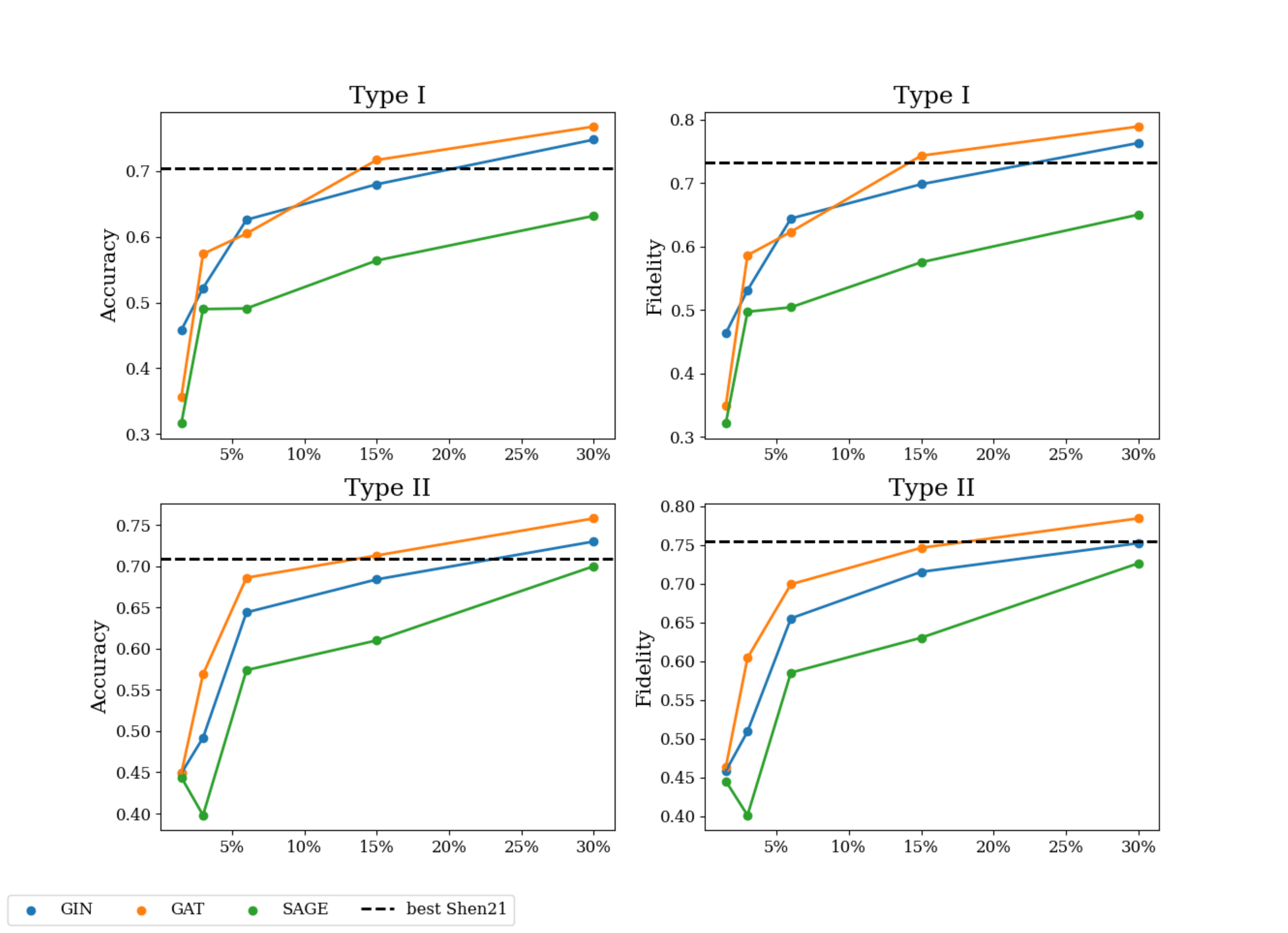} 
\caption{The average accuracies and fidelities on the Citeseer dataset and GIN target with projection response using different percentages of nodes in the stealing process. We compare with the highest accuracy and fidelity achieved by  \cite{shen2021model}'s best-performing surrogate model using 30\% of dataset nodes.}
\label{fig:query_projection}
\end{figure}
\vspace{1cm}

\begin{figure}[h!]
\centering
\includegraphics[trim={2.5cm 2cm 2cm 2.3cm},clip,width=0.7\linewidth]{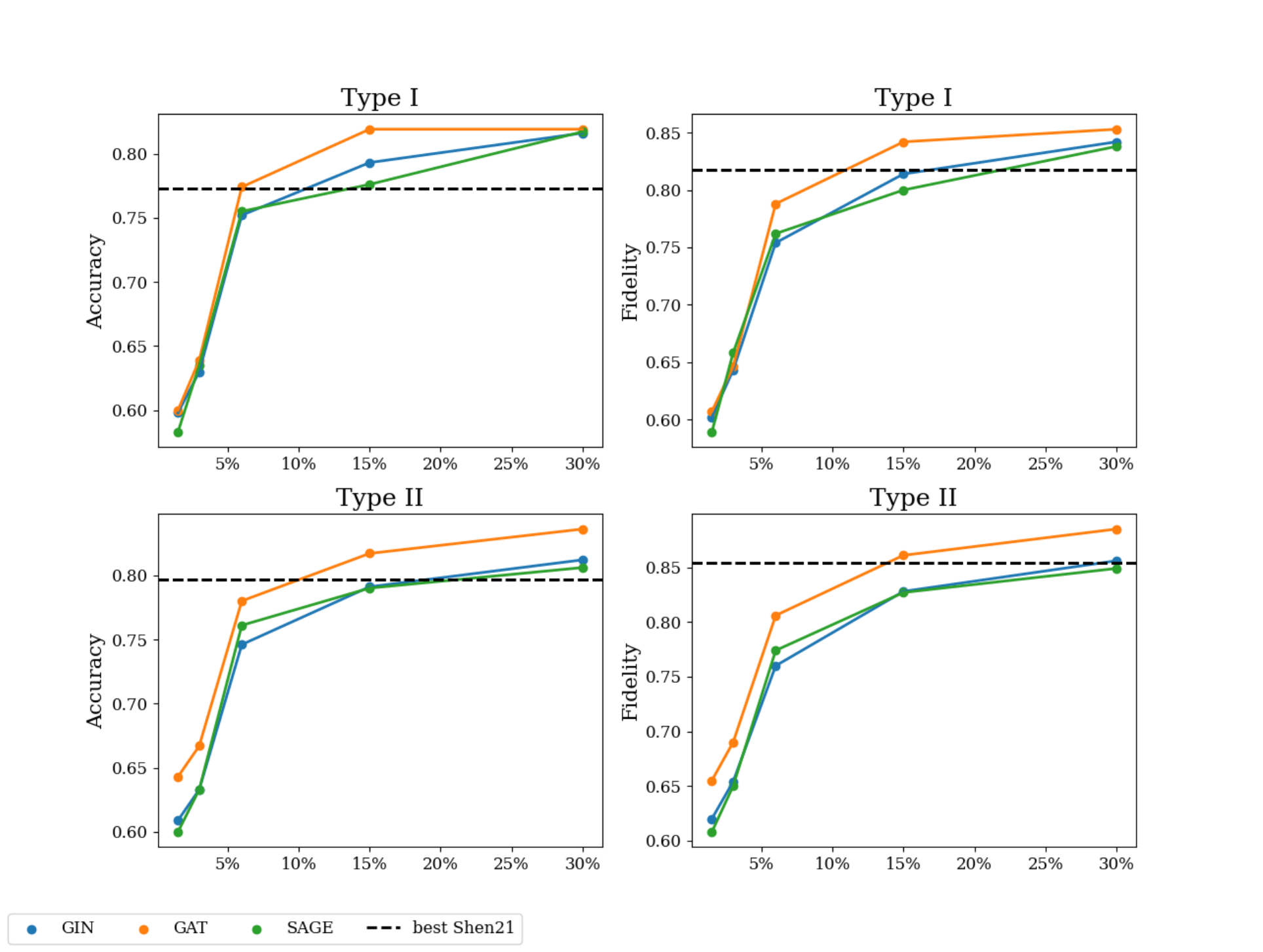} 
\includegraphics[trim={0 0cm 20cm 23cm},clip,width=0.5\linewidth]{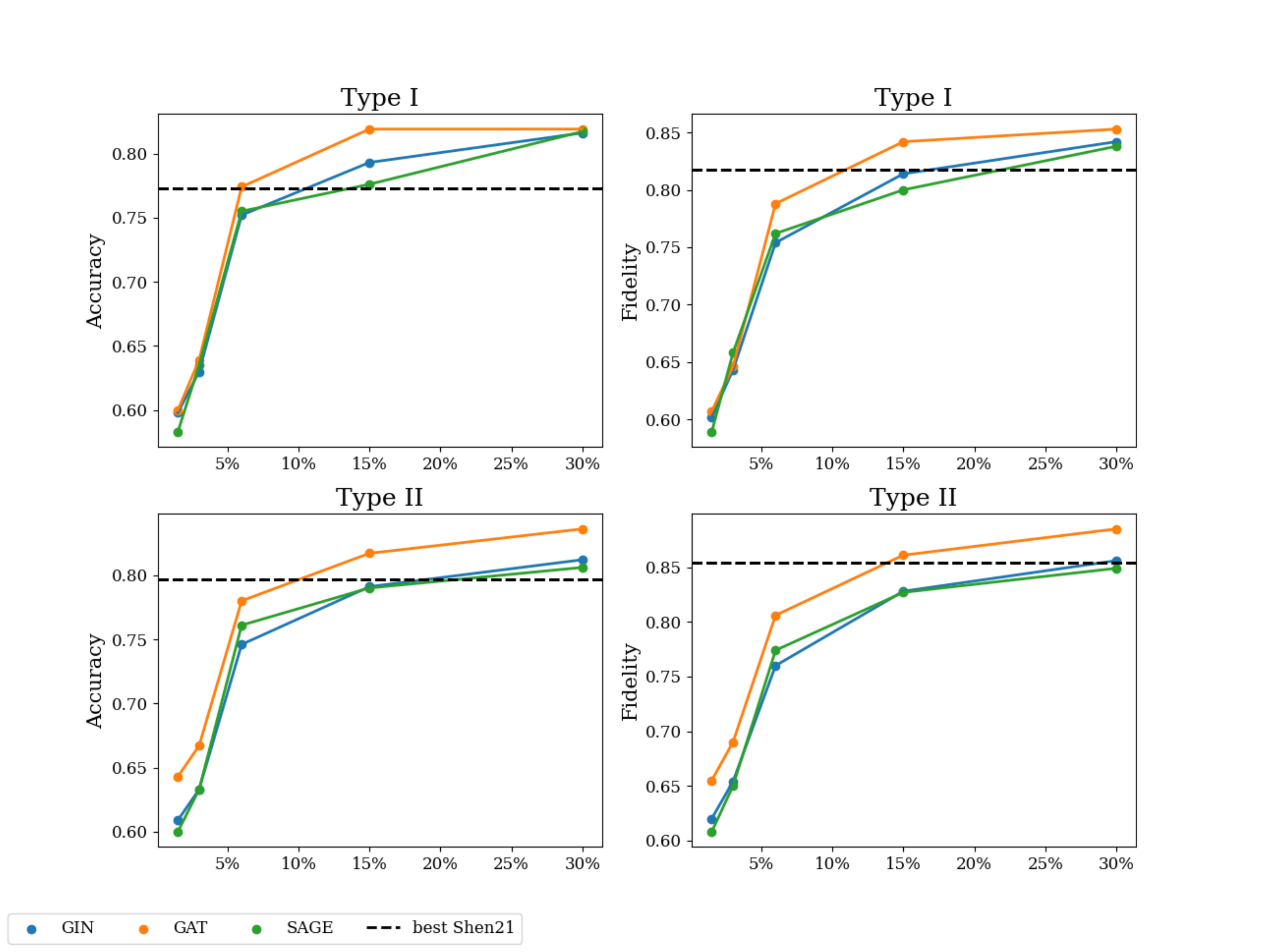} 
\caption{The average accuracy and fidelity on the Citeseer dataset and GIN target with prediction response using different percentages of nodes in the stealing process. We compare with the highest accuracy and fidelity achieved by  \cite{shen2021model}'s best-performing surrogate model using 30\% of dataset nodes.}
\label{fig:query_prediction}
\end{figure}

\clearpage
\section{Additional results - defense}\label{sec:appendix_defense}

\begin{figure}[!h]
\centering
\includegraphics[width=0.5\linewidth]{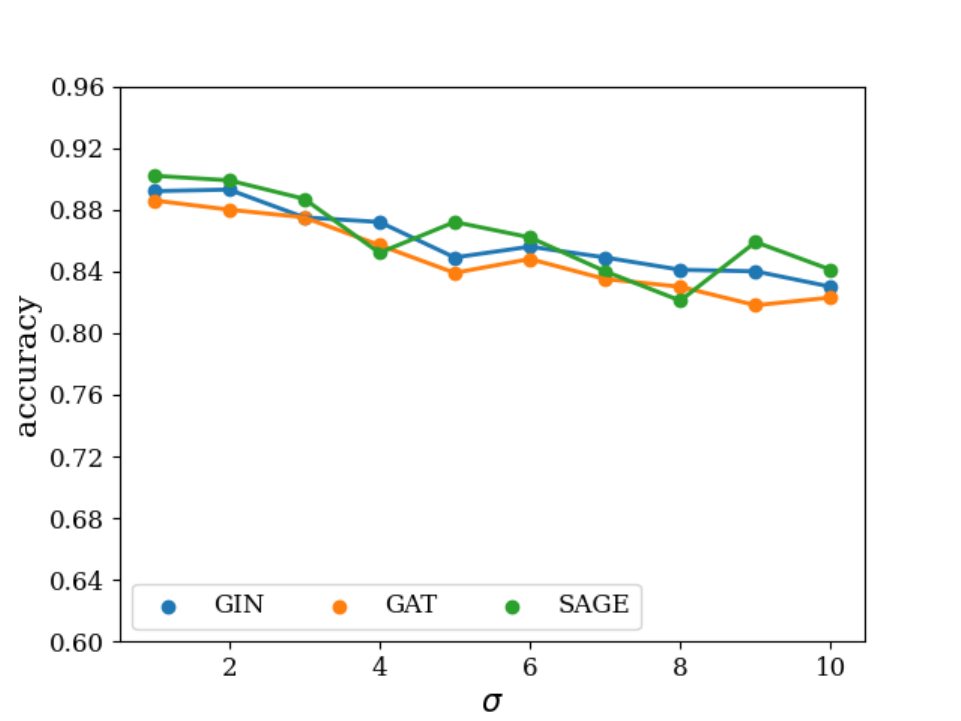} 
\caption{Accuracy scores after adding Gaussian noise with a standard deviation $\sigma$ to prediction response from target model GIN on the ACM dataset.}
\label{fig:defense_prediction}
\end{figure}

\begin{figure}[!h]
\centering
\includegraphics[width=0.5\linewidth]{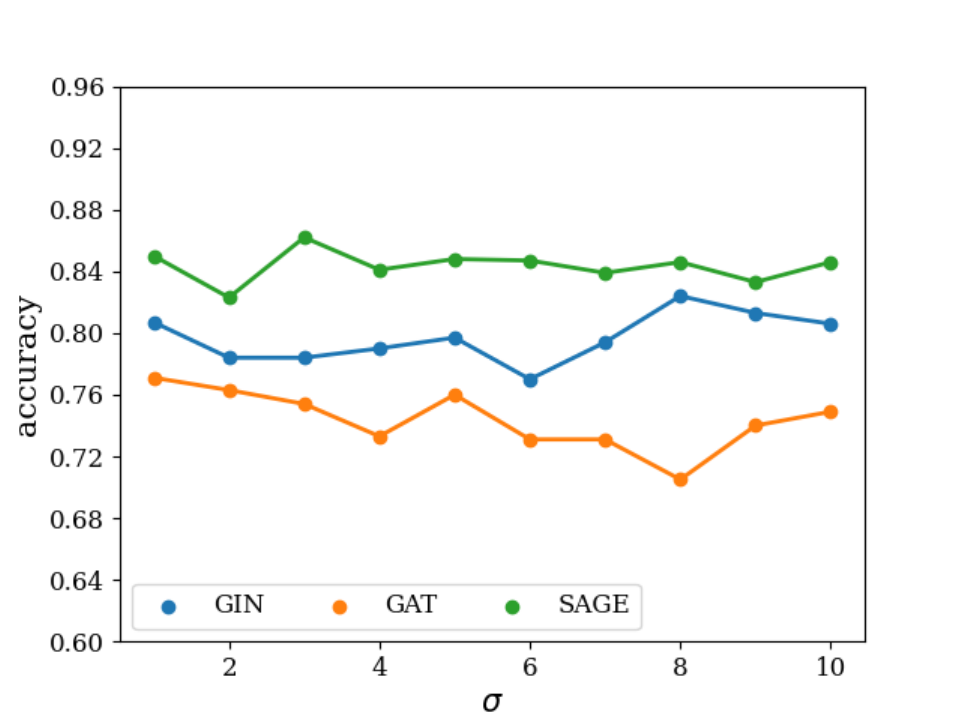} 
\caption{Accuracy scores after adding Gaussian noise with a standard deviation $\sigma$ to t-SNE projection response from target model GIN on the ACM dataset.}
\label{fig:defense_projection}
\end{figure}

\end{document}